\pdfoutput=1
\documentclass{article} 
\usepackage{iclr2024_conference,times}

\usepackage{amsmath,amsfonts,bm}

\def\eqref#1{equation~\ref{#1}}

\def\1{\bm{1}}

\DeclareMathAlphabet{\mathsfit}{\encodingdefault}{\sfdefault}{m}{sl}
\SetMathAlphabet{\mathsfit}{bold}{\encodingdefault}{\sfdefault}{bx}{n}

\iffalse 
    \newcommand\todo[1]{}

    \newcommand{\sihyun}[1]{}
    \newcommand{\weili}[1]{}
    \newcommand{\dean}[1]{}
    \newcommand{\boyi}[1]{}
    \newcommand{\anima}[1]{}
    \newcommand{\jinwoo}[1]{}
\else 
    \newcommand{\todo}[1]{{\textcolor{red}{[[TODO: {#1}]]}}}

    \newcommand{\sihyun}[1]{\textcolor{magenta}{[sihyun: {#1}]}}
    \newcommand{\weili}[1]{\textcolor{green}{[weili: {#1}]}}
    \newcommand{\dean}[1]{\textcolor{blue}{[de-an: {#1}]}}
    \newcommand{\boyi}[1]{\textcolor{orange}{[boyi: {#1}]}}
    \newcommand{\anima}[1]{\textcolor{purple}{[anima: {#1}]}}
    \newcommand{\jinwoo}[1]{\textcolor{navy}{[jinwoo: {#1}]}}
    
\fi

\usepackage{hyperref}
\usepackage{url}
\usepackage{xspace}
\usepackage{dsfont}
\usepackage{afterpage}

\usepackage{enumitem}
\usepackage{booktabs}
\usepackage{caption} 
\usepackage{multirow} 
\usepackage{enumitem}
\usepackage{colortbl} 

\usepackage{algorithm}
\usepackage{algpseudocode}                  
\usepackage{setspace}                       
\usepackage{lipsum} 
\usepackage{subcaption} 

\usepackage{mathtools} 
\usepackage{bm} 
\usepackage{soul} 
\usepackage{wrapfig}

\newcommand{\eg}{\emph{e.g.}} 
\newcommand{\ie}{\emph{i.e.}} 
\newcommand{\bc}{\mathbf{c}}
\newcommand{\bx}{\mathbf{x}}
\newcommand{\bz}{\mathbf{z}}
\newcommand{\bv}{\mathbf{v}}

\newcommand{\bu}{\mathbf{u}}
\newcommand{\stdv}[1]{\scriptsize$\pm$#1}

\newcommand{\lname}{content-motion latent diffusion model\xspace}
\newcommand{\sname}{CMD\xspace}

\definecolor{cornellred}{rgb}{0.7, 0.11, 0.11}
\definecolor{cadmiumgreen}{rgb}{0.0, 0.42, 0.24}
\definecolor{aliceblue}{rgb}{0.91, 0.94, 0.97}
\definecolor{darkblue}{rgb}{0.83, 0.89, 0.97}
\definecolor{Red7}{rgb}{0.941, 0.243, 0.243}
\definecolor{Green7}{RGB}{55, 178, 77}

\usepackage{pifont}
\newcommand{\cmark}{\ding{51}}%
\newcommand{\xmark}{\ding{55}}%
\newcommand{\ck}{\color{Green7}{\cmark}}
\newcommand{\xk}{\color{Red7}{\xmark}}
\sethlcolor{aliceblue}

\hypersetup{
  linkcolor = cornellred,
  citecolor  = cadmiumgreen,
  colorlinks = true,
}

\title{
Efficient Video Diffusion Models via 
\\ Content-Frame Motion-Latent Decomposition
}

\author{\hspace{-0.07in}Sihyun Yu$^{1}$\footnote[1]{} \,\,\, Weili Nie$^{2}$ \,\,\, De-An Huang$^{2}$ \,\,\, Boyi Li$^{2,3}$ \,\,\,Jinwoo Shin$^{1}$ \,\,\, Anima Anandkumar$^{4}$  \\
\hspace{-0.07in}$^{1}$KAIST \,\,\, $^{2}$NVIDIA Corporation \quad $^{3}$UC Berkeley \,\,\, $^{4}$Caltech \\
}

\iclrfinalcopy 
\begin{document}
\maketitle
\def\thefootnote{*}\footnotetext{Work done during an internship at NVIDIA. Project page: \url{https://sihyun.me/CMD}.}
\vspace{-0.15in}
\begin{abstract}
Video diffusion models have recently made great progress in generation quality, but are still limited by the high memory and computational requirements.
This is because current video diffusion models often attempt to process high-dimensional videos directly. 
To tackle this issue, we propose \lname (\sname), a novel, efficient extension of pretrained image diffusion models for video generation. 
Specifically, we propose an autoencoder that succinctly encodes a video as a combination of a content frame (like an image) and a low-dimensional motion latent representation. The former represents the common content, and the latter represents the underlying motion in the video, respectively.
We generate the content frame by fine-tuning a pretrained image diffusion model, and we generate the motion latent representation by training a new lightweight diffusion model. 
A key innovation here is the design of a compact latent space that can directly and efficiently utilize a pretrained image model, which has not been done in previous latent video diffusion models. 
This leads to considerably better quality generation and reduced computational costs. 
For instance, \sname can sample a video 7.7$\times$ faster than prior approaches by generating a video of 512$\times$1024 resolution and length 16 in 3.1 seconds. 
Moreover, \sname achieves an FVD score of 238.3 on WebVid-10M, 18.5\% better than the previous state-of-the-art of 292.4.
\end{abstract}

\vspace{-0.05in}
\vspace{-0.05in}
\section{Introduction}
\vspace{-0.05in}
\label{sec:intro}
\begin{wrapfigure}[24]{R}{0.5\textwidth}
\vspace{-0.17in}
\centering
\includegraphics[width=0.415\textwidth]{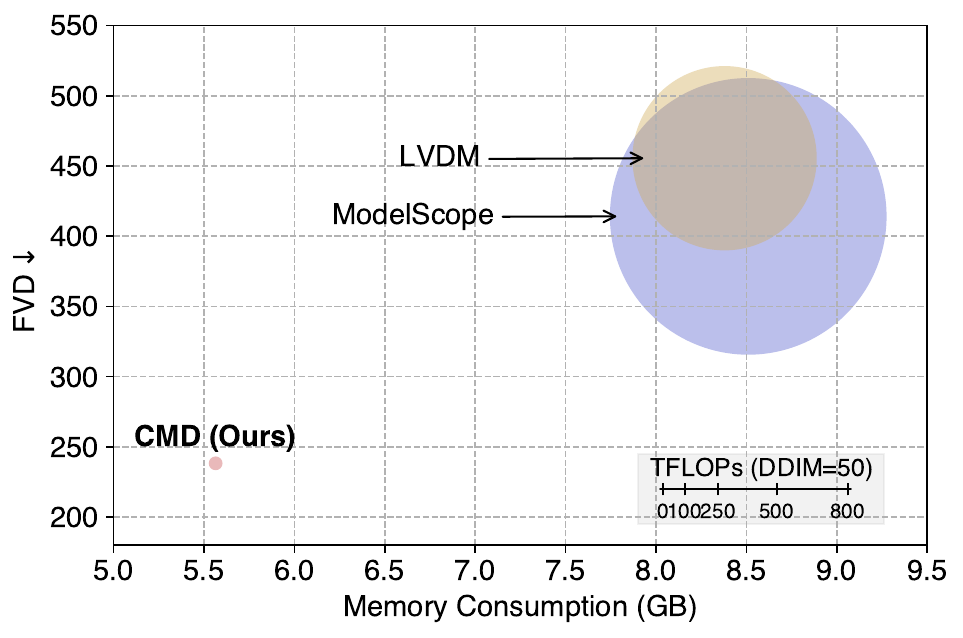}
\caption{Existing (text-to-)video diffusion models extended from image diffusion models often suffer from computation and memory inefficiency due to extremely high-dimensionality and temporal redundancy of video frames. Compared with these methods, \sname requires \textbf{$\sim$16.7$\times$ less computation with only $\sim$66\% GPU memory usage in sampling}, while achieving significantly better video generation quality. FLOPs and memory consumption are measured with a single NVIDIA A100 40GB GPU to generate a single video of a resolution 512$\times$1024 and length 16.}
\label{fig:teaser}
\end{wrapfigure}

Recently, deep generative models have exhibited remarkable success in synthesizing photo-realistic and high-resolution images using diffusion models (DMs) \citep{ho2021denoising,nichol2021improved,song2021scorebased,karras2022edm} and even achieving promising results in difficult text-to-image (T2I) generation \citep{rombach2021highresolution,saharia2022photorealistic,balaji2022ediffi}. Inspired by the success in the image domain, several works have focused on solving a considerably more challenging task of video generation. In particular, they have attempted to design DMs specialized for videos and shown reasonable video generation quality~\citep{ho2022video,yang2022diffusion,yu2023video,blattmann2023align}. Nevertheless, unlike the image domain, there is still a considerable gap in video quality between generated and real-world videos. This is mainly due to the difficulty of collecting a large training dataset of high-quality videos~\citep{ho2022video,ge2023preserve} and the high dimensionality of video data as cubic arrays, leading to a heavy memory and computational burden \citep{he2022lvdm,yu2023video}.

\begin{figure}[t!]
    \centering
    \includegraphics[width=\linewidth]{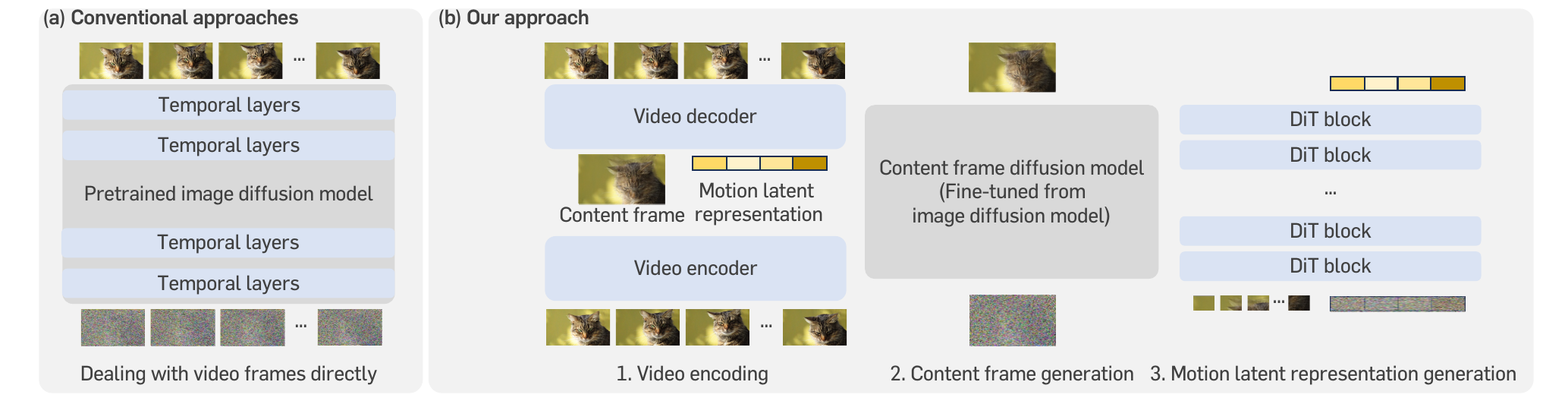}
    \caption{Comparison with (a) the conventional extension of image diffusion models for video generation and (b) our \sname. We mark the newly added parameters as \hl{blue}. Unlike common approaches that directly add temporal layers in pretrained image diffusion models for extension, \sname encodes each video as an image-like content frame and motion latents, and then fine-tunes a pretrained image diffusion model (\eg, Stable Diffusion~\citep{rombach2021highresolution}) for content frame generation and trains a new lightweight diffusion model (\eg, DiT~\citep{Peebles2022DiT}) for motion generation.     
    }
    \label{fig:concept}
    \vspace{-0.16in}
\end{figure}

To tackle the data collection issue, several video DM approaches leverage pretrained image DMs for video generation~\citep{he2022lvdm,singer2022make,luo2023videofusion,ge2023preserve}. Due to the rich visual knowledge already learned from image datasets, the use of image DMs in video generation leads to better generation quality and faster training convergence compared to training a video DM from scratch~\citep{an2023latent,blattmann2023align}. However, since these video models directly generate high-dimensional videos as cubic arrays, they still entail high memory consumption and computational costs, especially for high-resolution and long videos. 

Another line of video DM approaches focuses on alleviating memory and computational inefficiency by first projecting the video into a low-dimensional latent space and then training a DM in the latent space~\citep{yu2023video}. In particular, these approaches consider both the temporal coherency of videos as well as frame-wise compression in video encoding to obtain the maximum efficiency. However, such latent video DMs are only trained on a limited amount of video data and do not incorporate pretrained image models, which limits their video generation quality.

\textbf{Our approach.} We address the aforementioned shortcomings by introducing \lname (\sname), a memory- and computation-efficient latent video DM that leverages visual knowledge present in pretrained image DMs. \sname is a two-stage framework that first compresses videos to a succinct latent space and then learns the video distribution in this latent space. A key difference compared to existing latent video DMs is the design of a latent space that directly incorporates a pretrained image DM.  See Figure~\ref{fig:concept} for an illustration.

In the first stage, we learn a low-dimensional latent decomposition into a content frame (like an image) and latent motion representation through an autoencoder. Here, we design the content frame as a weighted sum of all frames in a video, where the weights are learned to represent the relative importance of each frame. In the second stage, we model the content frame distribution by fine-tuning a pretrained image DM without adding any new parameters. It allows \sname to leverage the rich visual knowledge in pretrained image DMs. In addition, we design a new lightweight DM to generate motion latent representation conditioned on the given content frame. Such designs avoid us having to deal directly with video arrays, and thus, one can achieve significantly better memory and computation efficiency than prior video DM approaches built on pretrained image DMs.

We highlight the main contributions of this paper below:
\vspace{-0.05in}
\begin{itemize}[leftmargin=*,itemsep=0mm]
    \item We propose an efficient latent video DM, termed \lname (\sname). 
    \item We validate the effectiveness of \sname on popular video generation benchmarks, including UCF-101~\citep{soomro2012ucf101} and WebVid-10M~\citep{Bain21}. For instance, measured with FVD~(\citealt{unterthiner2018towards}; lower is better), our method achieves {238.3} in text-to-video (T2V) generation on WebVid-10M, {18.5\%} better than the prior state-of-the-art of 292.4.
    \item We show the memory and computation efficiency of \sname. For instance, to generate a single video of resolution 512$\times$1024 and length 16, \sname only requires 5.56GB memory and 46.83 TFLOPs, while recent Modelscope~\citep{wang2023modelscope} requires 8.51GB memory and 938.9 TFLOPs, significantly larger than the requirements of \sname (see Figure \ref{fig:teaser}).
\end{itemize}

\begin{figure}[t!]
    \centering
    \includegraphics[width=\linewidth]{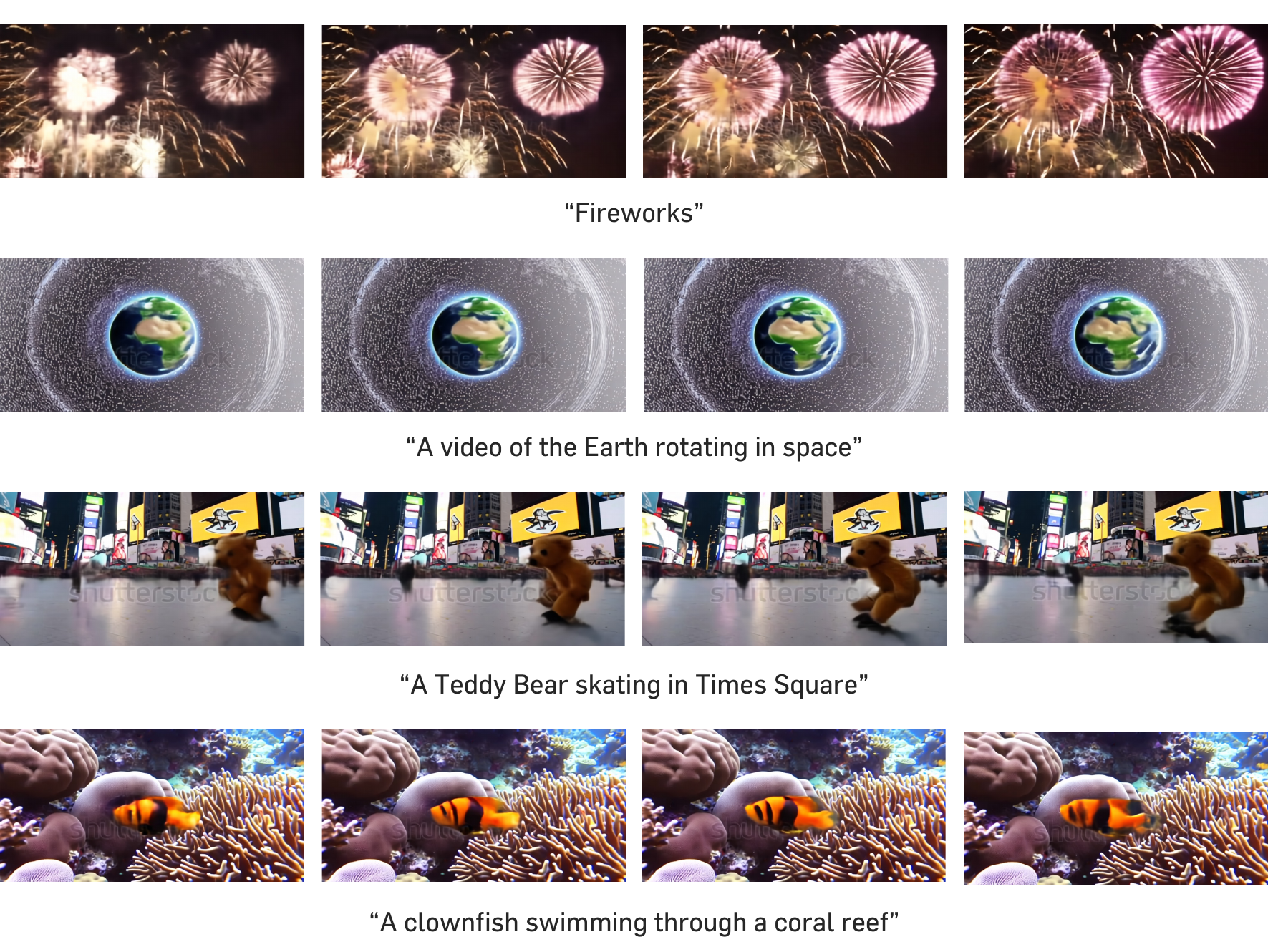}
    \caption{\textbf{512$\times$1024 resolution, 16-frame text-to-video generation results} from our \sname. We visualize video frames with a stride of 5. We provide more examples with different text prompts in Appendix~\ref{appen:more_qual}, as well as their illustrations as video file formats in the supplementary material.}
    \label{fig:main_qual}
\end{figure}

\section{Related Work}
\label{sec:related}
In this section, we provide a brief overview of some of the important relevant literature. For a more extensive discussion with a detailed explanation of other methods, see Appendix~\ref{appen:related}.

\textbf{Latent diffusion models.}
Diffusion models have suffered from memory and computation inefficiency because they require a large number of iterations in high-dimensional input space for sampling~\citep{ho2021denoising}. To mitigate this issue, several works have considered training diffusion models in a low-dimensional latent space, learned by an autoencoder \citep{zeng2022lion,xu2023geometric,ben2023ldm3d}. In particular, this approach has shown remarkable success in the image domain~\citep{rombach2021highresolution} to greatly improve efficiency as well as achieve high-quality synthesis results conditioned at a complex text prompt. Similarly, our work aims to design a latent diffusion model for videos~\citep{he2022lvdm,yu2023video} to alleviate the inefficiencies.

\noindent\textbf{Video generation.}
Numerous works have actively focused on solving the challenging problem of video synthesis. Previously, generative adversarial network (GAN;~\citealt{goodfellow2014generative}) based approaches \citep{gordon2021latent,tian2021good, fox2021stylevideogan,munoz2021temporal,yu2022digan,skorokhodov2021stylegan,singer2022make} were proposed to achieve the goal, mostly by extending popular image GAN architectures~\citep{karras2020analyzing}. Recently, there have been several works that encode videos as sequences of discrete tokens~\citep{van2017neural}, where they either generate tokens in an autoregressive manner \citep{kalchbrenner2017video,weissenborn2020scaling,rakhimov2020latent,yan2021videogpt,ge2022long} or a non-autoregressive manner \citep{yu2023magvit}. In addition, with the success of diffusion models~\citep{ho2021denoising,nichol2021improved} in image generation, recent methods exploit diffusion models for videos~\citep{ho2022video,harvey2022flexible,yang2022diffusion,hoppe2022diffusion,singer2022make,lu2023vdt}, achieving promising results in modeling complex video distribution. Inspired by their success, we also aim to build a new video diffusion model to achieve better video synthesis quality. 

\textbf{Text-to-video (T2V) generation.}
Following the success of text-to-image (T2I) generation~\citep{rombach2021highresolution,saharia2022photorealistic,balaji2022ediffi}, several works have attempted to solve a more challenging task of T2V generation. The main challenge of T2V generation is to resolve the tremendous training costs of diffusion models and the difficulty in collecting large-scale and high-quality video data. Predominant approaches \citep{ho2022imagen,wang2023videofactory,an2023latent,blattmann2023align,ge2023preserve,he2022lvdm,singer2022make} have achieved this by fine-tuning pretrained T2I diffusion models by adding temporal layers (\eg, temporal attention and 3D convolution layers) to the 2D U-Net architecture \citep{saharia2022photorealistic}. However, they suffer from high memory consumption and computational costs due to the unfavorable increase of input dimension in high-resolution and long videos. To tackle this issue, a few recent works have focused on alternative extension strategies that avoid dealing with entire raw video frames directly~\citep{luo2023videofusion}. Specifically, they achieve it by proposing frame-by-frame generation with an additional lightweight diffusion model. However, our extension is based on a latent diffusion model approach to encoding videos as content frames and motion latent representation to reduce the input dimension and learn video diffusion models on such compact latent representation. 
\section{CMD: Content-Motion Latent Diffusion Model}
\label{sec:method}

Consider a condition-video pair dataset $\mathcal{D}$, where each sample $(\bc, \bx^{1:L}) \in \mathcal{D}$ is drawn from an unknown data distribution $p_{\text{data}} (\bx^{1:L}, \bc)$. Here, each $\bc$ denotes a condition (\eg, video class or text description) of the corresponding $\bx^{1:L}$, and each $\bx^{1:L} \coloneqq (\bx^{1},\ldots,\bx^{L})$ is a video clip of length $L>1$ with a resolution $H \times W$, \ie, $\bx^{\ell} \in \mathbb{R}^{C \times H \times W}$ with a channel size $C$. Using $\mathcal{D}$, We aim to learn a conditional model distribution $p_{\text{model}}(\bx^{1:L}|\bc)$ to match the data distribution $p_{\text{data}} (\bx^{1:L} | \bc)$.

Our main idea is to encode each video into an ``image-like'' content frame and succinct motion latent representation, where pretrained image diffusion models can be used to generate content frames due to the similarity between natural images and content frames. By doing so, rich visual knowledge learned from image data is leveraged for video synthesis, leading to better generation quality as well as reduced training costs. Given content frames, the video generation task thus reduces to designing a motion diffusion model to generate much lower-dimensional motion latent representation.

In the rest of this section, we explain our \lname (\sname) in detail. In Section~\ref{subsec:ldm}, we provide an overview of diffusion models. In Section~\ref{subsec:framework}, we describe our video encoding scheme and design choices of diffusion models for video generation.

\subsection{Diffusion models}
\label{subsec:ldm}
The main concept of diffusion models is to learn the target distribution $p_{\text{data}}(\bx)$ via a gradual denoising process from Gaussian distribution $\mathcal{N}(\mathbf{0}_{\bx}, \mathbf{I}_{\bx})$ to $p_{\text{data}}(\bx)$. Specifically, diffusion models learn a \emph{reverse} process $p (\bx_{t-1} | \bx_{t})$ of the pre-defined \emph{forward} process $q(\bx_t | \bx_{0})$ that gradually adds the Gaussian noise starting from $p_{\text{data}}(\bx)$ for $1\leq t \leq T$ with a fixed $T>0$. Here, for $\bx_0 \sim p_{\text{data}}(\bx)$, $q(\bx_t | \bx_{t-1})$ can be formalized as $q(\bx_t| \bx_{t-1}) \coloneqq \mathcal{N}(\bx_t; \alpha_{t}\bx_0, \sigma_t^2\mathbf{I}_\bx)$, where $\sigma_t$ and $\alpha_t \coloneqq 1 - \sigma_t^{2}$ are pre-defined hyperparameters with $0 < \sigma_1 < \ldots < \sigma_{T-1} < \sigma_T =1$. If $T$ is sufficiently large, the reverse process $p (\bx_{t-1} | \bx_{t})$ can be also formalized as the following Gaussian distribution:
\begin{align}
    p(\bx_{t-1}|\bx_{t}) \coloneqq \mathcal{N}\Big(\bx_{t-1}; \frac{1}{\sqrt{\alpha_t}}\big(\bx_t - \frac{\sigma_t^2}{\sqrt{1-\bar{\alpha}_t}}\bm{\epsilon}_{\bm{\theta}}(\bx_t, t)\big), \sigma_t^2\mathbf{I}_{\bx}\Big),
\end{align}
where $\bar{\alpha}_t \coloneqq \prod_{i=1}^{t} (1-\sigma_i^2)$ for $1 \leq t \leq T$. Here, $\bm{\epsilon}_{\bm{\theta}}(\bx_t, t)$ can be trained as a denoising autoencoder parameterized by $\bm{\theta}$ using the $\bm{\epsilon}$-prediction objective with a noise $\bm{\epsilon} \sim \mathcal{N} (\mathbf{0}_{\bx}, \mathbf{I}_{\bx})$ \citep{ho2021denoising}:
\begin{align}
    \mathbb{E}_{\bx_0, \bm{\epsilon}, t} \Big[ ||\bm{\epsilon} - \bm{\epsilon}_{\bm{\theta}}(\bx_t, t)||_2^2 \Big]\,\, \text{where } \bx_t = \sqrt{\bar{\alpha}_t}\bx_0 + \sqrt{1-\bar{\alpha}_t}\bm{\epsilon}.
\end{align}
As the sampling process of diffusion models usually requires a large number of network evaluations $p(\bx_{t-1}|\bx_{t})$ (\eg, 1,000 in DDPM;~\citealt{ho2021denoising}), their generation cost becomes especially high if one defines diffusion models in the high-dimensional data space.
To mitigate this issue, several works have proposed latent diffusion models~\citep{rombach2021highresolution,he2022lvdm}: they train the diffusion model in a low-dimensional latent space that encodes the data, thus reducing the computation and memory cost. Inspired by their success, our work follows a similar idea of latent diffusion models to improve both training and sampling efficiency for video synthesis.

\begin{figure}[t!]
    \vspace{-0.1in}
    \centering    
    \includegraphics[width=.9\linewidth]{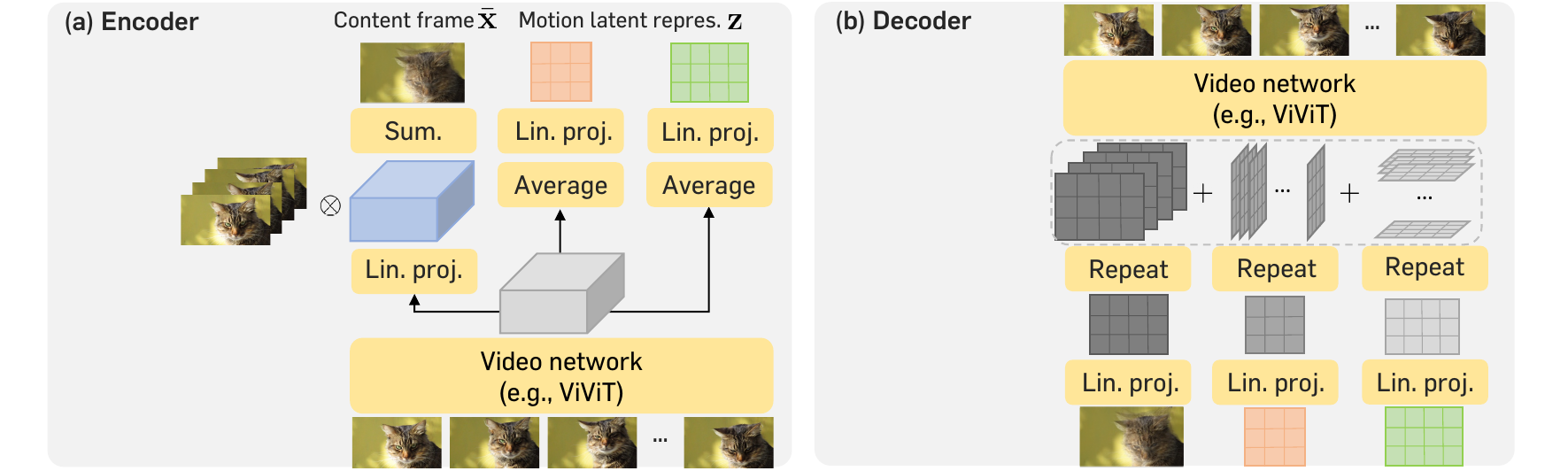}
    \vspace{-0.05in}
    \caption{\textbf{Illustration of our autoencoder}. Encoder: We compute relative importance of all frames (blue) for a content frame and motion latent representation. Decoder: Using the content frame and motion latent representation, we construct a cubic tensor for video network to reconstruct the video. 
    }
    \label{fig:autoenc}
    \vspace{-0.15in}
\end{figure}

\subsection{Efficient extension of image diffusion models for videos}
\label{subsec:framework}
\sname consists of an autoencoder and two latent diffusion models. First, we train an autoencoder that encodes a video $\bx^{1:L}$ as a single content frame $\bar{\bx}$ and low-dimensional motion latent representation $\bz$. After that, learning the target distribution $p_{\text{data}} (\bx^{1:L} | \bc)$ becomes to learn the following distribution: $p(\bar{\bx}, \bz|\bc)=p(\bz|\bar{\bx}, \bc)p(\bar{\bx}|\bc)$. We model each distribution through two diffusion models, where we utilize a pretrained image diffusion model for learning the content frame distribution $p(\bar{\bx}|\bc)$.

\textbf{Autoencoder.}
We train our autoencoder using a simple reconstruction objective (\eg, $\ell_2$ loss) to encode a video input $\bx^{1:L}$. We provide an illustration of the encoder and decoder in Figure~\ref{fig:autoenc}.

Our encoder $f_{\bm{\phi}}$ consists of a base network $f_{\bm{\phi}_{B}}$ and two heads $f_{\bm{\phi}_{I}}, f_{\bm{\phi}_{M}}$ for computing $\bar{\bx}, \bz$ (respectively) with a parameterization $\bm{\phi} \coloneqq ({\bm{\phi}_{B}},{\bm{\phi}_{I}},{\bm{\phi}_{M}})$. Here, the base network $f_{\bm{\phi}_{B}}: \mathbb{R}^{C \times L \times  H \times W} \to \mathbb{R}^{ C' \times L \times H' \times W'}$ (with $f_{\bm{\phi}_{B}}(\bx^{1:L}) = \bu$) maps a video $\bx^{1:L}$ to hidden feature $\bu$ with a channel size $C'$, where we adopt a video transformer (\eg, ViViT;~\citealt{arnab2021vivit}) as $f_{\bm{\phi}_{B}}$. Next, the head $f_{\bm{\phi}_{I}}: \mathbb{R}^{ C' \times L \times H' \times W'} \to \mathbb{R}^{C \times L \times  H \times W}$ returns relative importance among video frames $\bx^{1}, \ldots, \bx^{L}$ to compute the content frame $\bar{\bx}$. Specifically, we compute $\bar{\bx}$ using $f_{\bm{\phi}_I}$ as: 
\begin{align}
    \bar{\bx} \coloneqq \sum_{\ell=1}^{L}\Big(\bx^{\ell} \otimes \sigma\big(f_{\bm{\phi}_{I}}(\bu)\big)^{\ell}\Big),  
\end{align}
where $\otimes$ denotes an element-wise product and $\sigma(\cdot)$ is a softmax function across the temporal axis. Consequently, the content frame $\bar{\bx}$ has the same dimension with each frame and lies in the space of interpolating consecutive frames, thus looking very similar to them (see Figure~\ref{fig:keyframe}). 

For motion latent representation $\bz$, we design it as a concatenation of two latents, \ie, $\bz=(\bz_{\mathrm{x}},\bz_{\mathrm{y}})$ with $\bz_{\mathrm{x}}\in\mathbb{R}^{D \times L \times H'}$ and $\bz_{\mathrm{y}} \in \mathbb{R}^{D \times L \times W'}$, where $\bz_{\mathrm{x}}, \bz_{\mathrm{y}}$ are computed from $\bu$ using $f_{\bm{\phi}_M}$ as follows: 
\begin{align}
    (\bz_{\mathrm{x}},\bz_{\mathrm{y}}) \coloneqq \big( f_{\bm{\phi}_{M}}(\bar{\bu}_{\mathrm{x}}), f_{\bm{\phi}_{M}}(\bar{\bu}_{\mathrm{y}})\big). 
\end{align}
Here, $\bar{\bu}_{\mathrm{x}} \in \mathbb{R}^{C' \times L \times H'}, \bar{\bu}_{\mathrm{y}}\in \mathbb{R}^{C' \times L \times W'}$ are two projected tensors of $\bu$ by simply averaging across $\mathrm{x}$-axis and $\mathrm{y}$-axis, respectively, and $f_{\bm{\phi}_{M}}$ is a 1$\times$1 convolutional layer that maps an input tensor from a channel size $C'$ to $D$. Such a 2D-projection-based motion encoding is motivated by recent triplane video encoding~\citep{kim2022scalable,yu2023video} that project videos to each $\mathrm{x},\mathrm{y},\mathrm{t}$ axis.

Similarly, we design a decoder network $g_{\bm{\psi}}$ as two embedding layers $g_{\bm{\psi}_I}, g_{\bm{\psi}_M}$ for $\bar{\bx}, \bz$ (respectively) and a video network $g_{\bm{\psi}_B}$ that returns the reconstruction of $\bx^{1:L}$ from the outputs of $g_{\bm{\psi}_I}, g_{\bm{\psi}_M}$. Specifically, $g_{\bm{\psi}_I}, g_{\bm{\psi}_M}$ maps $\bar{\bx}, \bz$ to have the same channel size $C'$:
\begin{align}
\bv^\mathrm{t} \coloneqq g_{\bm{\psi}_I} (\bar{\bx}) \in  \mathbb{R}^{C'\times H' \times W'}, \,\,
\bv^\mathrm{x} \coloneqq g_{\bm{\psi}_M} (\bz_{\mathrm{x}}) \in  \mathbb{R}^{C'\times L \times H'},\,\, \bv^\mathrm{y}  \coloneqq g_{\bm{\psi}_M} (\bz_{\mathrm{y}}) \in  \mathbb{R}^{C'\times L \times W'},
\end{align}
where we denote $\bv^\mathrm{t} = [v_{hw}^{\mathrm{t}}],
\bv^\mathrm{x} = [v_{\ell h}^{\mathrm{x}}],
\bv^\mathrm{y} = [v_{\ell w}^{\mathrm{y}}]$ with 
$v_{hw}^{\mathrm{t}}, v_{\ell h}^{\mathrm{x}}, v_{\ell w}^{\mathrm{y}} \in \mathbb{R}^{C'}$ 
for $\ell \in [1, L], \, h \in [1, H'], \,  w \in [1, W']$. After that, we compute the input of a video network $g_{\bm{\psi}_B}$, denoted by $\bv = [v_{\ell hw}] \in \mathbb{R}^{C' \times L \times H' \times W'}$, by taking the sum of the corresponding vectors of each $\bv^{\mathrm{t}}, \bv^{\mathrm{x}}, \bv^{\mathrm{y}}$, namely: 
\begin{align}
    v_{\ell hw} = v_{hw}^{\mathrm{t}} + v_{\ell h}^{\mathrm{x}} + v_{\ell w}^{\mathrm{y}} \,\, \text{for}\,\, 1\leq \ell \leq L,\,\,  1\leq h \leq H',\,\,  1\leq w \leq W',
\end{align}
and then $\bv$ is passed to $g_{\bm{\psi}_B}: \mathbb{R}^{C' \times L \times H' \times W'} \to \mathbb{R}^{C \times L \times H \times W}$ to reconstruct the input video $\bx^{1:L}$. For $g_{\bm{\psi}_B}$, we use the same video transformer architecture as $f_{\bm{\phi}_B}$.

\textbf{Content frame diffusion model.}
Recall that the content frame $\bar{\bx}$ is computed as a weighted sum of video frames $\bx^{1},\ldots,\bx^{L}$ and thus it resembles natural images. Hence, for training the content frame diffusion model to learn $p(\bar{\bx}|\bc)$, we directly fine-tune the pretrained image diffusion model $\bm{\epsilon}_{\bm{\theta}_I} (\bx_t, \bc, t)$ without adding additional parameters. In particular, we use condition-content-frame pairs $(\bc, \bar{\bx})$ computed from the dataset $\mathcal{D}$ and use the denoising objective for fine-tuning: 
\begin{align}
    \mathbb{E}_{\bar{\bx}_0, \bm{\epsilon}, t} \Big[ ||\bm{\epsilon} - \bm{\epsilon}_{\bm{\theta}_I} (\bar{\bx}_t, \bc, t)||_2^2 \Big]\,\, \text{where } \bar{\bx}_t = \sqrt{\bar{\alpha}_t}\bar{\bx}_0 + \sqrt{1-\bar{\alpha}_t}\bm{\epsilon}.
\end{align}

Note that this fine-tuning is memory-efficient since it does not increase input dimension, and it can be trained efficiently due to the small gap between content frames and natural images.

\textbf{Motion diffusion model.} To learn the conditional distribution $p(\bz|\bar{\bx}, \bc)$, we train a lightweight diffusion model  $\bm{\epsilon}_{\bm{\theta}_M} (\bz_t, \bc, \bar{\bx}, t)$. For the network architecture, we exploit DiT~\citep{Peebles2022DiT}, a recently proposed Vision Transformer (ViT) backbone \citep{dosovitskiy2020image} for diffusion models, due to its better performance and efficiency. Accordingly, for a denoising target $\bz_t$, we pass it to the model as a sequence of patch embeddings. Next, for an input condition $\bc$, we follow the same conditioning scheme of the original DiT that passes it through the AdaIN layers~\citep{huang2017arbitrary}. For the conditioning content frame $\bar{\bx}$, rather than passing it through the AdaIN layers, we feed it as input-level patch embeddings like $\bz_t$ to provide ``dense conditions'' to the model for predicting motion latent representation $\bz$ (see Figure~\ref{fig:concept}). Using these inputs, we train the model via the denoising objective:
\begin{align}
    \mathbb{E}_{{\bz}_0, \bm{\epsilon}, t} \Big[ ||\bm{\epsilon} - \bm{\epsilon}_{\bm{\theta}_M} (\bz_t, \bc, \bar{\bx}, t)||_2^2 \Big]\,\, \text{where } \bz_t = \sqrt{\bar{\alpha}_t}\bz_0 + \sqrt{1-\bar{\alpha}_t}\bm{\epsilon}.
\end{align}
We observe that a lightweight model can quickly converge to well-predicting motion latent representation $\bz$, mainly due to two factors: (a) the rich information provided by the conditions ($\bc, \bar{\bx}$), and (b) the low dimensionality of motion latent representation $\bz$. Moreover, one can use a larger patch size for $\bar{\bx}$ (condition) than $\bz$ (prediction target) to reduce the total sequence length of input patches to the DiT network, thus further decreasing the computational cost (see Section~\ref{subsec:analysis}).
\section{Experiments}
\label{sec:exp}
In Section~\ref{subsec:exp_setup}, we provide setups for our experiments. In Section~\ref{subsec:main_exp}, we present the main results, including qualitative results of visualizing generated videos. Finally, in Section~\ref{subsec:analysis}, we conduct extensive analysis to validate the effect of each component as well as to show the efficiency of \sname in various aspects, compared with previous text-to-video generation methods. 

\vspace{-0.02in}
\subsection{Setups}
\vspace{-0.02in}
\label{subsec:exp_setup}

\textbf{Datasets.} 
We mainly consider UCF-101~\citep{soomro2012ucf101} and WebVid-10M~\citep{Bain21} for the evaluation. We also use MSR-VTT~\citep{xu2016msr} for a zero-shot evaluation of the text-to-video models. For model training, we use only train split and exclude test (or validation) sets for all datasets. We provide more details, including how they are preprocessed in Appendix~\ref{appen:datasets}.

\textbf{Baselines.}
For class-conditional (non-zero-shot) generation on UCF-101, we consider recent DIGAN \citep{yu2022digan}, TATS \citep{ge2022long}, CogVideo \citep{hong2023cogvideo}, Make-A-Video \citep{singer2022make}, and MAGVIT \citep{yu2023magvit} as baselines. For zero-shot evaluations, we compare with recent CogVideo, LVDM~\citep{he2022lvdm}, ModelScope~\citep{wang2023modelscope}, VideoLDM~\citep{blattmann2023align}, VideoFactory~\citep{wang2023videofactory}, PYoCo~\citep{ge2023preserve}, GODIVA~\citep{wu2021godiva}, and N\"UWA~\citep{wu2022nuwa}. See Appendix~\ref{appen:baselines} for more details. 

\textbf{Training details.}
In all experiments, videos are clipped to 16 frames for both training and evaluation. For a video autoencoder, we use TimeSFormer~\citep{bertasius2021space} as a backbone. For the content frame diffusion model, we use pretrained Stable Diffusion (SD) 1.5 and 2.1-base~\citep{rombach2021highresolution}, where each video frame is first encoded by SD image autoencoder into a latent frame with an 8$\times$ downsampling ratio and output channel size $C=4$.
For the motion diffusion model, we use DiT-L/2 (for UCF-101) and DiT-XL/2 (for WebVid-10M) as in the original DiT paper~\citep{Peebles2022DiT}, where ``L'' and ``XL'' specify the model sizes and ``2'' denotes patch size of 2$\times$2 when converting input into a sequence of patches. We provide all other details in Appendix~\ref{appen:training}.

\textbf{Metrics.}
Following the experimental setup in recent representative video generation literature~\citep{skorokhodov2021stylegan,yu2023magvit}, we mainly use Fr\'echet video distance (FVD;~\citealt{unterthiner2018towards}, lower is better) for evaluation. To measure text-video alignment, we additionally measure CLIPSIM (\citealt{wu2021godiva}, higher is better) and compare the values with the baselines. We provide more details of evaluation metrics and how they are computed in Appendix~\ref{appen:metrics}.

\begin{figure}[t]
\vspace{-0.07in}
\small\centering
\begin{minipage}[t]{0.48\textwidth}
\centering\small
\captionof{table}{\textbf{Class-conditional video generation on UCF-101.} \# denotes the model also uses the test split for training. $\downarrow$ indicates lower values are better. Bolds indicate the best results, and we mark our method by blue. We mark the method by * if the score is evaluated with 10,000 real data and generated videos, otherwise we use 2,048 videos. For a zero-shot setup, we report the dataset size used for training.}
\begin{tabular}{l c c} 
\toprule
{Method} & {Zero-shot} & {FVD $\downarrow$}  \\
     \midrule
    {DIGAN$^{\text{\#}}$}     
    & {No} & {465\stdv{12}}  \\
    {TATS}     
    & {No} & {332\stdv{18}}  \\
    {CogVideo}     
    & {No} & {305}  \\
    {VideoFusion}     
    & {No} & {173}  \\
    \midrule
    \rowcolor{aliceblue}
    {\textbf{\sname (Ours)}} 
    & {\textbf{No}} & {\textbf{107}\stdv{9}}  \\ 
    \midrule
    {Make-A-Video*}   
    & {No} & {367} \\
    {MAGVIT*}     
    & {No} & {76\stdv{2}} \\
    \midrule
    \rowcolor{aliceblue}
    {\textbf{\sname (Ours)}*} 
    & {\textbf{No}} & {\textbf{73\stdv{2}}}  \\ 
    \midrule
    {{VideoFactory}} 
    & {{Yes (130M)}} & {{410}} \\
    {{PYoCo}} 
    & {{Yes (22.5M)}} & {{\textbf{355}}} \\
    \midrule
    {CogVideo}     
    & {Yes (5.4M)} & {702}  \\
    {LVDM}         
    & {Yes (10.7M)} & {642}\\
    {ModelScope}   
    & {Yes (10.7M)} & {640}\\
    {VideoLDM}     
    & {Yes (10.7M)} & {551} \\
    {VideoGen}     
    & {Yes (10.7M)} & {554} \\
    \midrule
    \rowcolor{aliceblue}
    {\textbf{\sname (Ours)}} 
    & {\textbf{Yes (10.7M)}} & {\textbf{504}}  \\ 
    \bottomrule
\end{tabular}
\label{tab:ucf101}
\end{minipage}
~
\begin{minipage}[t]{0.48\textwidth}
\centering
\captionof{table}{\textbf{T2V generation on MSR-VTT.} $\uparrow$ indicates higher scores are better. Bolds indicate the best results, and we mark our method by blue. We report the dataset size. * denotes LAION-5B~\citep{schuhmann2022laion} is jointly used. 
}
\vspace{-0.03in}
\begin{tabular}{l c c c}
\toprule
{Method} & {Zero-shot} & {CLIPSIM $\uparrow$} \\
\midrule
    {GODIVA}       & {No} & {0.2402} \\
    {N\"UWA}         & {No} & {0.2409} \\
    \midrule
    {{VideoFactory}} & {{Yes (130M)}} & {{0.3005}} \\
    {{Make-A-Video}} & {{Yes (100M)}} & {{\textbf{0.3049}}} \\
    \midrule
    {CogVideo}   & {Yes (5.4M)} & {0.2631} \\
    {LVDM}         & {Yes (10.7M)} & {0.2381} \\
    {VideoLDM}     & {Yes (10.7M)} & {0.2929} \\
    {ModelScope*}   & {Yes (10.7M)} & {\textbf{0.2930}} \\
    \midrule
    \rowcolor{aliceblue}
    \textbf{\sname (Ours)} & {\textbf{Yes (10.7M)}} & 0.2894 \\
    \bottomrule
\label{tab:msrvtt}
\end{tabular}
\vspace{-0.2in}
\captionof{table}{\textbf{T2V generation on WebVid-10M.} $\downarrow$ and $\uparrow$ indicate lower and higher scores are better, respectively. Bolds indicate the best results, and we mark our method by blue. {cfg denotes classifier-free guidance scale.}}
\vspace{0.055in}
\begin{tabular}{l c c}
\toprule
{Method} & {FVD $\downarrow$} & {CLIPSIM $\uparrow$} \\
\midrule
    {LVDM}         & {455.5} & {0.2751} \\
    {ModelScope}   & {414.1} & {0.3000} \\
    {VideoFactory} & {292.4} & {\textbf{0.3070}} \\
    \midrule    \rowcolor{aliceblue}
    \textbf{\sname (Ours); cfg=9.0} & {{\textbf{238.3}}} & {{0.3020}} \\
    \bottomrule
\label{tab:webvid}
\end{tabular}
\end{minipage}
\vspace{-0.3in}
\end{figure}

\vspace{-0.03in}
\subsection{Main results}
\vspace{-0.03in}
\label{subsec:main_exp}

\textbf{Qualitative results.}
We visualize several text-to-video generation results from \sname in Figure~\ref{fig:main_qual}.
As shown in this figure, generated videos contain the detailed motion and contents provided by text prompts and achieve great temporal coherency, leading to a smooth video transition. In particular, the background is preserved well between different video frames in the generated video with the prompt. For instance, ``A Teddy Bear Skating in Times Square'' maintains details of Times Square well across different video frames. Note that each frame has a resolution of 512$\times$1024, where we achieve such a high-resolution video generation without requiring any spatiotemporal upsamplers. We provide more qualitative results with other text prompts in Appendix~\ref{appen:more_qual}.

\textbf{Quantitative results.}
Table~\ref{tab:ucf101} provides the non-zero-shot generation result on UCF-101 by training all models from scratch on UCF-101 (including the content frame diffusion model). As shown in this table, \sname outperforms all other video generation methods, indicating our framework design itself is an effective video generation method irrespective of the exploitation of pretrained image diffusion models. Moreover, we consider text-to-video generation by training \sname on WebVid-10M with the pretrained SD backbone fine-tuned for content frame generation. As shown in Table~\ref{tab:ucf101}~and~\ref{tab:webvid}, our model shows better FVD scores than previous approaches if the same amount of data is used. Moreover, our model achieves comparable or even better CLIPSIM scores, compared with state-of-the-art as shown in Table~\ref{tab:msrvtt} and \ref{tab:webvid}, indicating a good text-video alignment. \sname shows a slightly worse CLIPSIM score than ModelScope and VideoLDM on MSR-VTT, but note that our model (1.6B) is $\sim$1.9$\times$ smaller than VideoLDM (3.1B). Moreover, ModelScope \emph{jointly} trains on 5 billion image-text pairs along with video data to avoid catastrophic forgetting, in contrast to \sname that does not use any image data for training once provided pretrained image diffusion models.

\begin{figure}[t!]
    \centering
    \begin{subfigure}[t]{0.3\textwidth}
    \includegraphics[width=\linewidth]{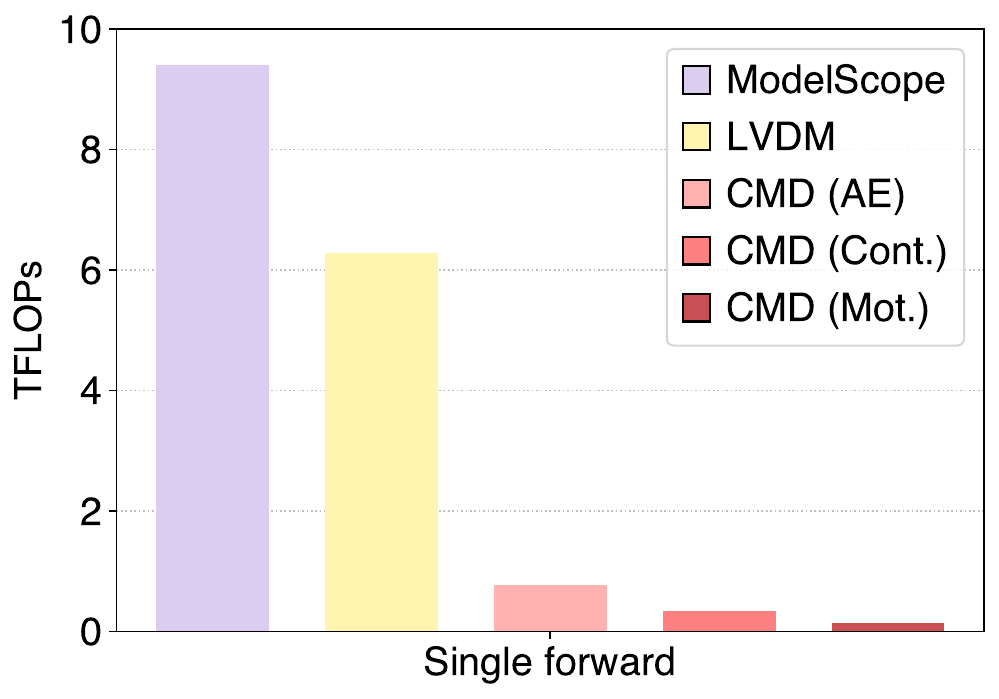}        
    \caption{FLOPs}
    \label{subfig:training_flops}
    \end{subfigure}
    \begin{subfigure}[t]{0.3\textwidth}
    \includegraphics[width=\linewidth]{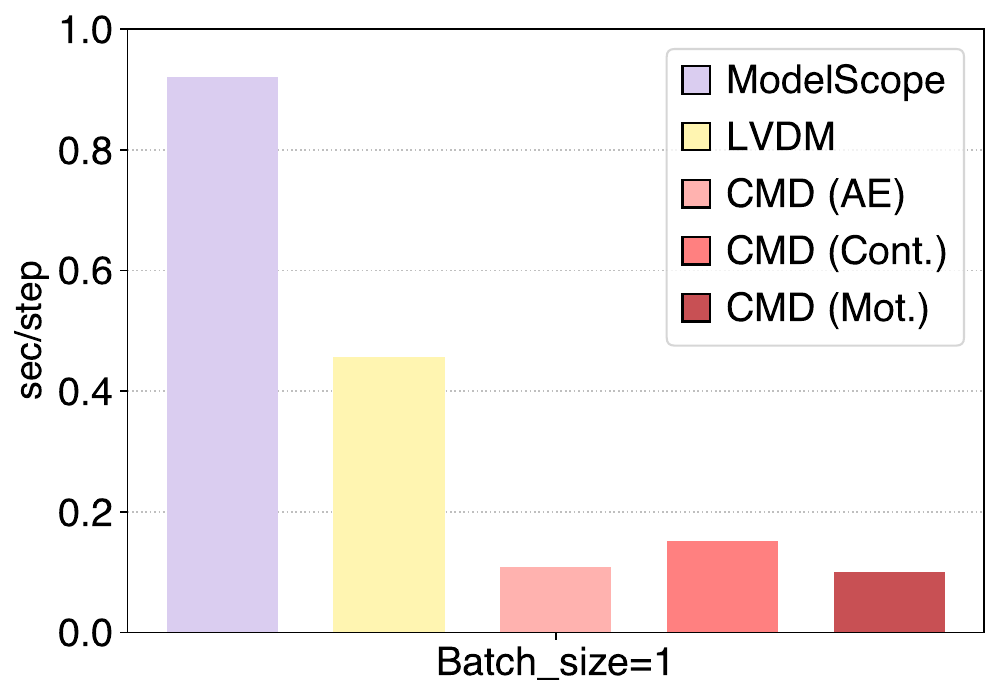}        
    \caption{Time}
    \label{subfig:training_time}
    \end{subfigure}
    \begin{subfigure}[t]{0.3\textwidth}
    \includegraphics[width=\linewidth]{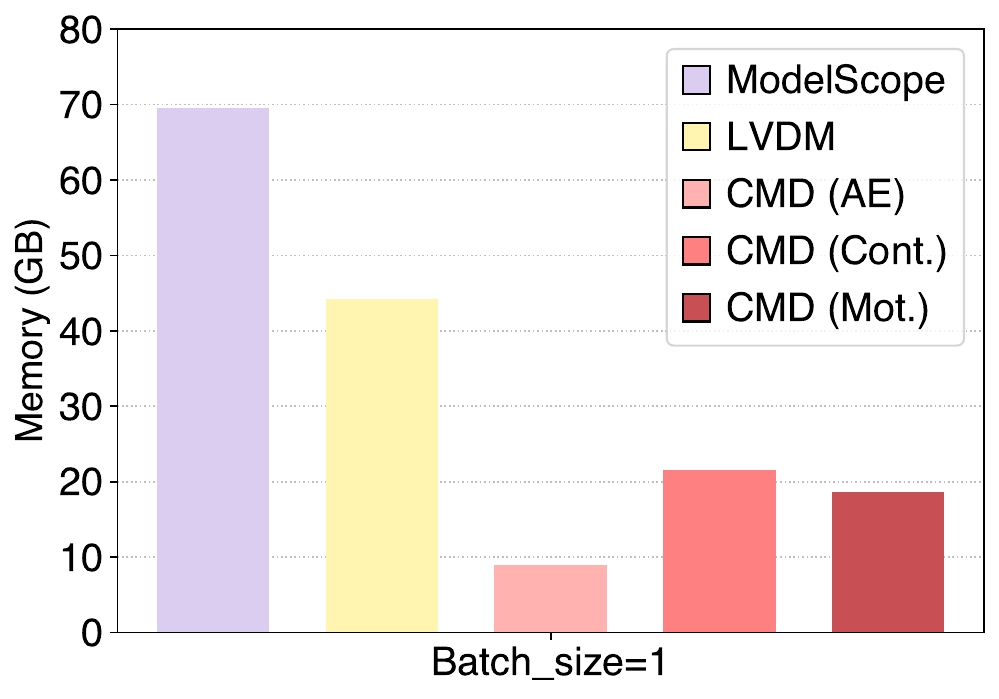}        
    \caption{Memory}
    \label{subfig:training_mem}
    \end{subfigure}
    \vspace{-0.05in}
    \caption{\textbf{Training efficiency.} (a) FLOPs, (b) sec/step, and (c) memory (GB) of different methods that are trained on 16-frame videos with resolution of $512\times512$ and batch size of 1. All values are measured with a single NVIDIA A100 80GB GPU with mixed precision. For a fair comparison, we do not apply gradient checkpointing for all models. See Appendix~\ref{appen:efficiency} for details.
    }
    \label{fig:component_efficiency}
\vspace{-0.1in}
\end{figure}
\begin{figure}[t!]
    \centering
    \begin{subfigure}[t]{0.3\textwidth}
    \includegraphics[width=\linewidth]{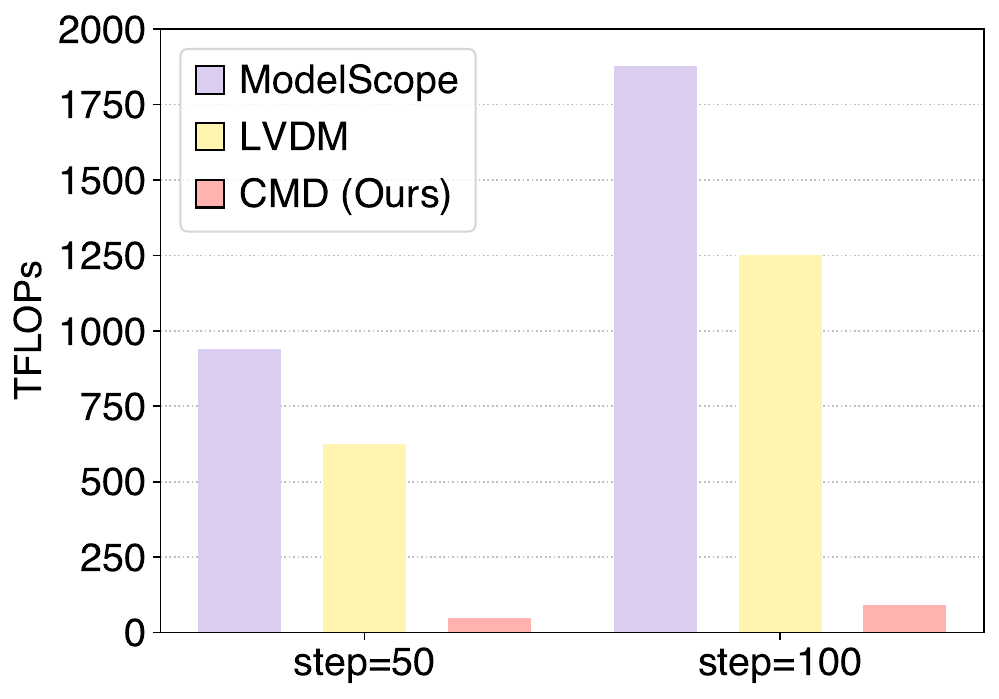}        
    \caption{FLOPs}
    \label{subfig:sampling_flops}
    \end{subfigure}
    \begin{subfigure}[t]{0.3\textwidth}
    \includegraphics[width=\linewidth]{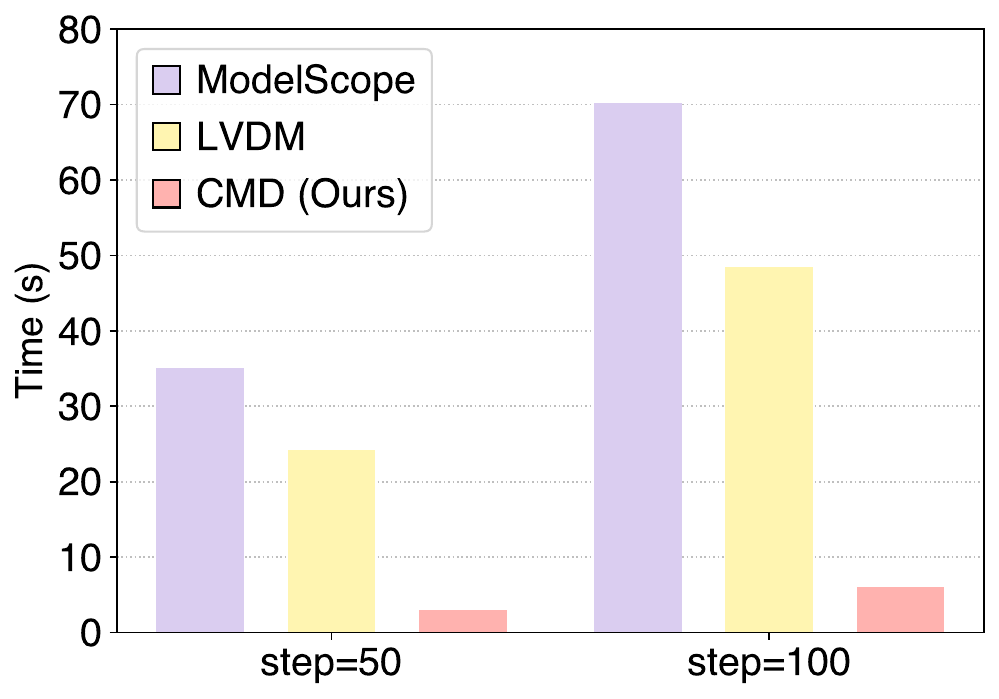}        
    \caption{Time}
    \label{subfig:sampling_time}
    \end{subfigure}
    \begin{subfigure}[t]{0.3\textwidth}
    \includegraphics[width=\linewidth]{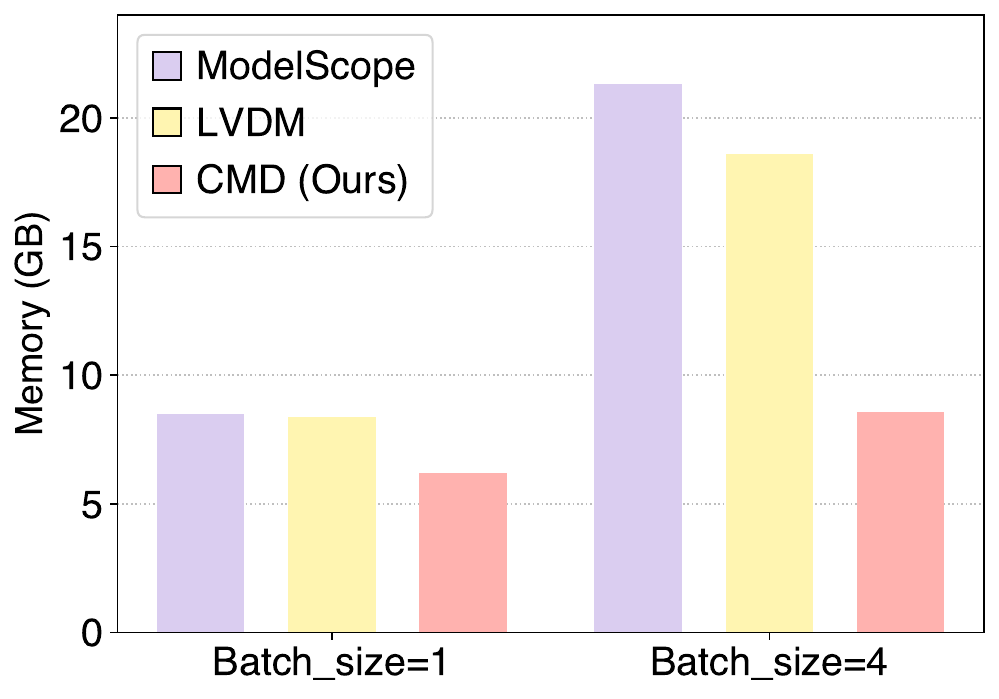}        
    \caption{Memory}
    \label{subfig:sampling_mem}
    \end{subfigure}
\vspace{-0.05in}
    \caption{\textbf{Sampling efficiency.} (a) FLOPs, (b) time (s), and (c) memory (GB) of different methods that sample a 16-frame video with resolution of $512\times1024$ (\ie, batch size = 1 by default). All values are measured with a single NVIDIA A100 40GB GPU with mixed precision. 
    Note that we exclude the cost of Stable Diffusion decoder for all measurements. See Appendix~\ref{appen:efficiency} for details.
    }
    \label{fig:sampling_efficiency}
    \vspace{-0.2in}
\end{figure}

\subsection{Analysis}
\label{subsec:analysis}
\textbf{Training efficiency.}
Figure~\ref{fig:component_efficiency} summarizes the computation (floating point operations; FLOPs), time, and memory consumption in training each component of \sname and compares the values with other public text-to-video diffusion models. As shown in these plots, all components of \sname require less memory and computation for training due to the decomposition of videos as two low-dimensional latent variables (content frame and motion latent representation). Notably, \sname shows significantly fewer FLOPs than prior methods: the bottleneck is in the autoencoder (0.77 TFLOPs) and is $\sim$12$\times$ more efficient than 9.41 TFLOPs of ModelScope. Note that if one sums up the FLOPs or training time of all three components in \sname, they are still significantly better than existing text-to-video diffusion models. We also note that the training of content frame diffusion models and motion diffusion models can be done in parallel. Thus, the training efficiency (in terms of time) can be further boosted. We also provide a model parameter size comparison in Appendix~\ref{appen:efficiency}. 

\textbf{Sampling efficiency.}
Figure~\ref{fig:sampling_efficiency} reports FLOPs, time, and memory consumption to sample videos. As shown in Figure~\ref{subfig:sampling_flops}, existing text-to-video diffusion models require tremendous computations for sampling since they directly input videos as high-dimensional cubic arrays. In particular, they overlook common contents in video frames (\eg, static background), and accordingly, many spatial layer operations (\eg, 2D convolutions) become unfavorably redundant and tremendous. However, \sname avoids dealing with giant cubic arrays, and thus, redundant operations are significantly reduced, resulting in a computation-efficient video generation framework. The sampling efficiency is also reflected in sampling time (Figure~\ref{subfig:sampling_time}); \sname only requires $\sim$3 seconds with a DDIM sampler~\citep{song2021denoising} using 50 steps, which is 10$\times$ faster than existing text-to-video diffusion models.

Not only improving computation efficiency, our method also exhibits great memory efficiency compared with existing methods due to the significantly reduced input dimension. Note that the improvement becomes more significant if the models sample multiple videos at once (\ie, a batch size larger than 1) because, in that case, the memory bottleneck mainly stems from the computation of intermediate features for sampling rather than the memory allocation of the model parameters. For instance, as shown in Figure~\ref{subfig:sampling_mem}, our model uses about 8.6GB GPU memory to generate 4 videos in parallel, 2.5$\times$ less consumption than the recent ModelScope model that requires more than 20GB.  

\textbf{Ablation studies.}
In Table~\ref{subtab:all}, we report the FVD values by using only some of the components in \sname. As shown in this table, each module in \sname shows reasonable performance, which validates our design choices for the overall framework. Moreover, in Table~\ref{subtab:ae}, we analyze the performance of the autoencoder under various setups; one can observe that the use of weighted sum in content frame design helps to achieve better reconstruction, and our autoencoder can encode videos with a longer length than 16 (\eg, $L=$24) with reasonable quality as well. Finally, Table~\ref{subtab:motion} shows that motion diffusion models exhibit a reasonable performance with large patch sizes, so one can control the tradeoff between computation efficiency and memory efficiency by adjusting the patch size.

\begin{table*}[t!]
\centering
\caption{
\textbf{Ablation studies.}
(a) FVD on UCF-101 to evaluate each component. Reconstruction: FVD between real videos and their reconstructions from our autoencoder. Motion prediction: FVD between real videos and predicted videos with the motion diffusion model conditioning on (ground-truth) content frames encoded by our autoencoder. Content generation: performance of \sname, where content frames are generated by our content frame diffusion model. (b) R-FVD of autoencoders on WebVid-10M with different channel sizes $D$, video lengths $L$, and the usage of weighted sum or not. (c) FVD of motion diffusion models on UCF-101 
with different content frame patch sizes.}
\vspace{-0.03in}
    \begin{subtable}[t]{0.32\linewidth}    
    \caption{Performance of each component}
    \centering\small
    \begin{tabular}{l c}
    \toprule
    {Task} & FVD \\
    \midrule
    Reconstruction & {7.72} \\
    Motion prediction  & {19.5}\\
    Content generation & {73.1}\\
    \bottomrule
    \end{tabular}
    \label{subtab:all}
    \end{subtable}    
    \begin{subtable}[t]{0.32\linewidth}    
    \centering\small
    \caption{Autoencoder}
    \begin{tabular}{c c c c}
    \toprule
    {$D$}  & {$L$} & {Weight.} & {R-FVD} \\ 
    \midrule 
    {16}   & {16}  & {\ck} & {56.8} \\
    {8}    & {16}  & {\ck} & {69.5} \\
    {8}    & {16}  & {\xk} & {76.1} \\
    {8}    & {24}  & {\ck} & {81.3} \\
    \bottomrule
    \label{subtab:ae}
    \end{tabular}        
    \end{subtable}
    \begin{subtable}[t]{0.32\linewidth}    
    \vspace{-0.14in}
    \centering\small
`    \caption{Motion diffusion}
    \begin{tabular}{l c c}
    \toprule
    {Config.}  & {$\bar{\bx}$ patch.} & {FVD} \\ 
    \midrule 
    {DiT-L/2} & 16 & 40.4 \\
    {DiT-L/2} & 8 & 32.9 \\
    {DiT-L/2} & 4 & 19.5 \\
    \bottomrule
    \label{subtab:motion}
    \end{tabular}        
    \end{subtable}    
\label{tab:ablation}
\vspace{-0.15in}
\end{table*}
\begin{figure}[t!]
    \centering
    \includegraphics[width=\linewidth]{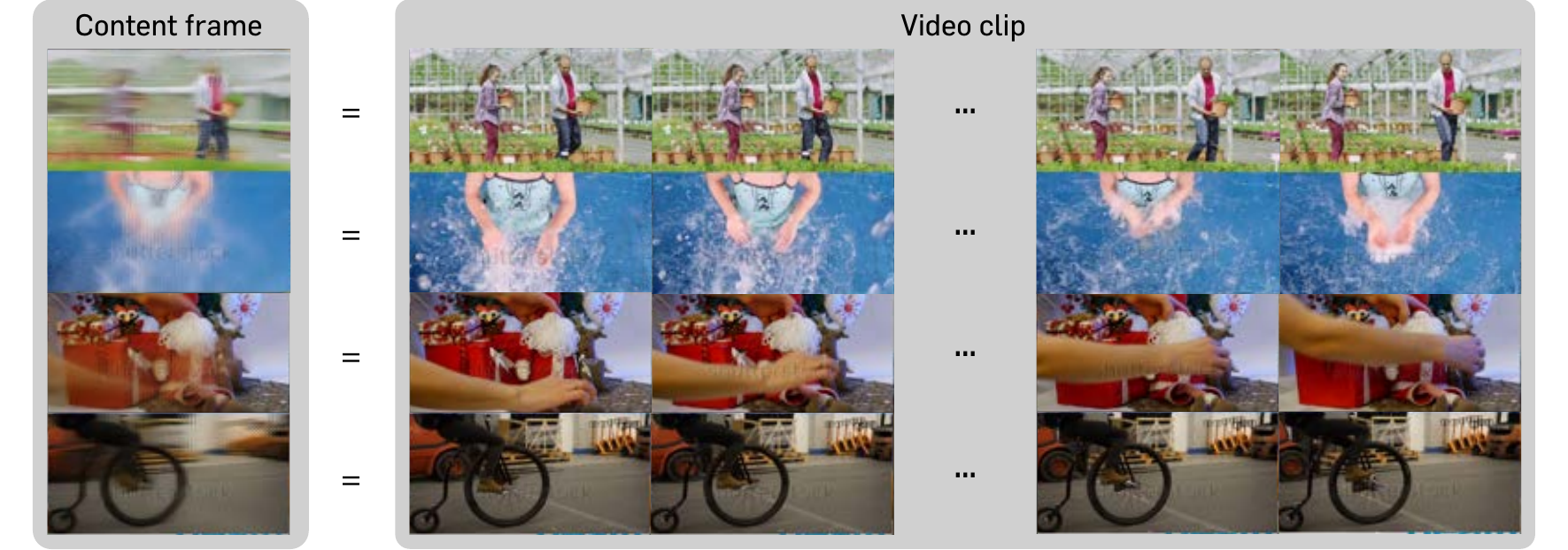}
    \caption{\textbf{Content frame visualization} with the corresponding video frames.
    }
    \label{fig:keyframe}
\vspace{-0.15in}
\end{figure}

\textbf{Content frame visualization.}
 Figure~\ref{fig:keyframe} visualizes videos in WebVid-10M and the corresponding content frames. As shown in this figure, the content frames resemble the original video frames, \ie, in the content frames, the background (\eg, buildings) and objects (\eg, a bicycle) appear similarly to the video frames. Moreover, one can observe that only the region with moving objects is corrupted, \eg, for the content frame of a video with a moving arm, an area where the arm appears is corrupted.

\section{Conclusion}
\label{sec:conclusion}
We proposed \sname, an efficient extension scheme of the image diffusion model for video generation. Our key idea is based on proposing a new encoding scheme that represents videos as content frames and succinct motion latents to improve computation and memory efficiency.
We hope our method will facilitate lots of intriguing directions for efficient large-scale video generation methods.

\textbf{Limitation and future work.}
In this work, we primarily focused on generating a video of a fixed length (\eg, $L=16$). One of the interesting future directions would be extending our method for long video synthesis, similar to PVDM~\citep{yu2023video} which considers clip-by-clip generation. Another interesting direction is to develop a better form of content frame and motion latents to encode video with higher quality but still enable exploiting pretrained image diffusion models. We provide a more detailed discussion of limitation and future work in Appendix~\ref{appen:future}.

\section*{Ethics statement}
We believe \sname can provide a positive impact in real-world scenarios related to content-creation applications. Since \sname can instantly synthesize videos from arbitrary user text prompts, it can save time for designers~\citep{villegas2023phenaki} who want to generate new content by providing them with an initial shape of such desired result. Moreover, given that the success of large text-to-image generation models \citep{rombach2021highresolution,saharia2022photorealistic,balaji2022ediffi} has facilitated intriguing applications such as image editing~\citep{brooks2023instructpix2pix, kim2023collaborative,meng2022sdedit} and personalized generation~\citep{ruiz2023dreambooth,gal2023an}, we expect developing a large-scale video generation framework will promote similar applications in the video domain as well~\citep{molad2023dreamix}.

In contrast, there also exists some potential negative impact of developing a large-scale generation framework to generate sensitive and malicious content, \eg, DeepFake~\citep{guera2018deepfake}, as discussed by some recent large-scale video generation works~\citep{villegas2023phenaki}. Although generated videos from \sname are relatively short and the frame quality is yet distinguishable from real-world videos, one should be aware of this issue and keep considering to develop a safe video generation framework in the future.

\section*{Reproducibility statement}
We provide implementation details (\eg, hyperparameter, model, and optimizer) and experiment setups (\eg, how the metrics are computed) in Section~\ref{sec:exp} and Appendix~\ref{appen:implementation}.

\section*{Acknowledgements}
SY thanks Subin Kim, Jaehyun Nam, Jihoon Tack, and anonymous reviewers for their helpful feedbacks on the early version of the manuscript. SY also acknowledges Seung Wook Kim for helping text-to-video model training.

\bibliography{iclr2024_conference}

\begin{thebibliography}{94}
\providecommand{\natexlab}[1]{#1}
\providecommand{\url}[1]{\texttt{#1}}
\expandafter\ifx\csname urlstyle\endcsname\relax
  \providecommand{\doi}[1]{doi: #1}\else
  \providecommand{\doi}{doi: \begingroup \urlstyle{rm}\Url}\fi

\bibitem[An et~al.(2023)An, Zhang, Yang, Gupta, Huang, Luo, and
  Yin]{an2023latent}
Jie An, Songyang Zhang, Harry Yang, Sonal Gupta, Jia-Bin Huang, Jiebo Luo, and
  Xi~Yin.
\newblock Latent-shift: Latent diffusion with temporal shift for efficient
  text-to-video generation.
\newblock \emph{arXiv preprint arXiv:2304.08477}, 2023.

\bibitem[Arnab et~al.(2021)Arnab, Dehghani, Heigold, Sun, Lu{\v{c}}i{\'c}, and
  Schmid]{arnab2021vivit}
Anurag Arnab, Mostafa Dehghani, Georg Heigold, Chen Sun, Mario Lu{\v{c}}i{\'c},
  and Cordelia Schmid.
\newblock Vivit: A video vision transformer.
\newblock In \emph{IEEE International Conference on Computer Vision}, 2021.

\bibitem[Babaeizadeh et~al.(2018)Babaeizadeh, Finn, Erhan, Campbell, and
  Levine]{babaeizadeh2018stochastic}
Mohammad Babaeizadeh, Chelsea Finn, Dumitru Erhan, Roy~H Campbell, and Sergey
  Levine.
\newblock Stochastic variational video prediction.
\newblock In \emph{International Conference on Learning Representations}, 2018.

\bibitem[Bain et~al.(2021)Bain, Nagrani, Varol, and Zisserman]{Bain21}
Max Bain, Arsha Nagrani, G{\"u}l Varol, and Andrew Zisserman.
\newblock Frozen in time: A joint video and image encoder for end-to-end
  retrieval.
\newblock In \emph{IEEE International Conference on Computer Vision}, 2021.

\bibitem[Balaji et~al.(2022)Balaji, Nah, Huang, Vahdat, Song, Kreis, Aittala,
  Aila, Laine, Catanzaro, et~al.]{balaji2022ediffi}
Yogesh Balaji, Seungjun Nah, Xun Huang, Arash Vahdat, Jiaming Song, Karsten
  Kreis, Miika Aittala, Timo Aila, Samuli Laine, Bryan Catanzaro, et~al.
\newblock ediffi: Text-to-image diffusion models with an ensemble of expert
  denoisers.
\newblock \emph{arXiv preprint arXiv:2211.01324}, 2022.

\bibitem[Ben Melech~Stan et~al.(2023)Ben Melech~Stan, Wofk, Fox, Redden,
  Saxton, Yu, Aflalo, Tseng, Nonato, Muller, et~al.]{ben2023ldm3d}
Gabriela Ben Melech~Stan, Diana Wofk, Scottie Fox, Alex Redden, Will Saxton,
  Jean Yu, Estelle Aflalo, Shao-Yen Tseng, Fabio Nonato, Matthias Muller,
  et~al.
\newblock {LDM3D}: Latent diffusion model for 3d.
\newblock \emph{arXiv e-prints}, 2023.

\bibitem[Bertasius et~al.(2021)Bertasius, Wang, and
  Torresani]{bertasius2021space}
Gedas Bertasius, Heng Wang, and Lorenzo Torresani.
\newblock Is space-time attention all you need for video understanding?
\newblock In \emph{International Conference on Machine Learning}, 2021.

\bibitem[Blattmann et~al.(2023)Blattmann, Rombach, Ling, Dockhorn, Kim, Fidler,
  and Kreis]{blattmann2023align}
Andreas Blattmann, Robin Rombach, Huan Ling, Tim Dockhorn, Seung~Wook Kim,
  Sanja Fidler, and Karsten Kreis.
\newblock Align your latents: High-resolution video synthesis with latent
  diffusion models.
\newblock In \emph{IEEE Conference on Computer Vision and Pattern Recognition},
  2023.

\bibitem[Brooks et~al.(2023)Brooks, Holynski, and
  Efros]{brooks2023instructpix2pix}
Tim Brooks, Aleksander Holynski, and Alexei~A Efros.
\newblock Instructpix2pix: Learning to follow image editing instructions.
\newblock In \emph{IEEE Conference on Computer Vision and Pattern Recognition},
  2023.

\bibitem[Carreira \& Zisserman(2017)Carreira and Zisserman]{carreira2017quo}
Joao Carreira and Andrew Zisserman.
\newblock Quo vadis, action recognition? a new model and the kinetics dataset.
\newblock In \emph{IEEE Conference on Computer Vision and Pattern Recognition},
  2017.

\bibitem[Chang et~al.(2022)Chang, Zhang, Jiang, Liu, and
  Freeman]{chang2022maskgit}
Huiwen Chang, Han Zhang, Lu~Jiang, Ce~Liu, and William~T Freeman.
\newblock Maskgit: Masked generative image transformer.
\newblock In \emph{IEEE Conference on Computer Vision and Pattern Recognition},
  2022.

\bibitem[Denton \& Birodkar(2017)Denton and Birodkar]{denton2017unsupervised}
Emily Denton and Vighnesh Birodkar.
\newblock Unsupervised learning of disentangled representations from video.
\newblock In \emph{Advances in Neural Information Processing Systems}, 2017.

\bibitem[Denton \& Fergus(2018)Denton and Fergus]{denton2018stochastic}
Emily Denton and Rob Fergus.
\newblock Stochastic video generation with a learned prior.
\newblock In \emph{International Conference on Machine Learning}, 2018.

\bibitem[Dosovitskiy et~al.(2020)Dosovitskiy, Beyer, Kolesnikov, Weissenborn,
  Zhai, Unterthiner, Dehghani, Minderer, Heigold, Gelly,
  et~al.]{dosovitskiy2020image}
Alexey Dosovitskiy, Lucas Beyer, Alexander Kolesnikov, Dirk Weissenborn,
  Xiaohua Zhai, Thomas Unterthiner, Mostafa Dehghani, Matthias Minderer, Georg
  Heigold, Sylvain Gelly, et~al.
\newblock An image is worth 16x16 words: Transformers for image recognition at
  scale.
\newblock In \emph{International Conference on Learning Representations}, 2020.

\bibitem[Esser et~al.(2021)Esser, Rombach, and Ommer]{esser2020taming}
Patrick Esser, Robin Rombach, and Björn Ommer.
\newblock Taming transformers for high-resolution image synthesis.
\newblock In \emph{IEEE Conference on Computer Vision and Pattern Recognition},
  2021.

\bibitem[Finn et~al.(2016)Finn, Goodfellow, and Levine]{finn2016unsupervised}
Chelsea Finn, Ian Goodfellow, and Sergey Levine.
\newblock Unsupervised learning for physical interaction through video
  prediction.
\newblock In \emph{Advances in Neural Information Processing Systems}, 2016.

\bibitem[Fox et~al.(2021)Fox, Tewari, Elgharib, and
  Theobalt]{fox2021stylevideogan}
Gereon Fox, Ayush Tewari, Mohamed Elgharib, and Christian Theobalt.
\newblock {StyleVideoGAN}: A temporal generative model using a pretrained
  {StyleGAN}.
\newblock \emph{arXiv preprint arXiv:2107.07224}, 2021.

\bibitem[Franceschi et~al.(2020)Franceschi, Delasalles, Chen, Lamprier, and
  Gallinari]{franceschi2020stochastic}
Jean-Yves Franceschi, Edouard Delasalles, Micka{\"e}l Chen, Sylvain Lamprier,
  and Patrick Gallinari.
\newblock Stochastic latent residual video prediction.
\newblock In \emph{International Conference on Machine Learning}, 2020.

\bibitem[Gal et~al.(2023)Gal, Alaluf, Atzmon, Patashnik, Bermano, Chechik, and
  Cohen-or]{gal2023an}
Rinon Gal, Yuval Alaluf, Yuval Atzmon, Or~Patashnik, Amit~Haim Bermano, Gal
  Chechik, and Daniel Cohen-or.
\newblock An image is worth one word: Personalizing text-to-image generation
  using textual inversion.
\newblock In \emph{International Conference on Learning Representations}, 2023.

\bibitem[Ge et~al.(2022)Ge, Hayes, Yang, Yin, Pang, Jacobs, Huang, and
  Parikh]{ge2022long}
Songwei Ge, Thomas Hayes, Harry Yang, Xi~Yin, Guan Pang, David Jacobs, Jia-Bin
  Huang, and Devi Parikh.
\newblock Long video generation with time-agnostic {VQGAN} and time-sensitive
  transformer.
\newblock In \emph{European Conference on Computer Vision}, 2022.

\bibitem[Ge et~al.(2023)Ge, Nah, Liu, Poon, Tao, Catanzaro, Jacobs, Huang, Liu,
  and Balaji]{ge2023preserve}
Songwei Ge, Seungjun Nah, Guilin Liu, Tyler Poon, Andrew Tao, Bryan Catanzaro,
  David Jacobs, Jia-Bin Huang, Ming-Yu Liu, and Yogesh Balaji.
\newblock Preserve your own correlation: A noise prior for video diffusion
  models.
\newblock In \emph{IEEE International Conference on Computer Vision}, 2023.

\bibitem[Goodfellow et~al.(2014)Goodfellow, Pouget-Abadie, Mirza, Xu,
  Warde-Farley, Ozair, Courville, and Bengio]{goodfellow2014generative}
Ian Goodfellow, Jean Pouget-Abadie, Mehdi Mirza, Bing Xu, David Warde-Farley,
  Sherjil Ozair, Aaron Courville, and Yoshua Bengio.
\newblock Generative adversarial nets.
\newblock In \emph{Advances in Neural Information Processing Systems}, 2014.

\bibitem[Gordon \& Parde(2021)Gordon and Parde]{gordon2021latent}
Cade Gordon and Natalie Parde.
\newblock Latent neural differential equations for video generation.
\newblock In \emph{NeurIPS 2020 Workshop on Pre-registration in Machine
  Learning}, 2021.

\bibitem[Guera \& Delp(2018)Guera and Delp]{guera2018deepfake}
David Guera and Edward~J. Delp.
\newblock Deepfake video detection using recurrent neural networks.
\newblock \emph{IEEE International Conference on Advanced Video and Signal
  Based Surveillance}, 2018.

\bibitem[Guo et~al.(2023)Guo, Yang, Rao, Wang, Qiao, Lin, and
  Dai]{guo2023animatediff}
Yuwei Guo, Ceyuan Yang, Anyi Rao, Yaohui Wang, Yu~Qiao, Dahua Lin, and Bo~Dai.
\newblock Animatediff: Animate your personalized text-to-image diffusion models
  without specific tuning.
\newblock \emph{arXiv preprint arXiv:2307.04725}, 2023.

\bibitem[Harvey et~al.(2022)Harvey, Naderiparizi, Masrani, Weilbach, and
  Wood]{harvey2022flexible}
William Harvey, Saeid Naderiparizi, Vaden Masrani, Christian Weilbach, and
  Frank Wood.
\newblock Flexible diffusion modeling of long videos.
\newblock In \emph{Advances in Neural Information Processing Systems}, 2022.

\bibitem[He et~al.(2022)He, Yang, Zhang, Shan, and Chen]{he2022lvdm}
Yingqing He, Tianyu Yang, Yong Zhang, Ying Shan, and Qifeng Chen.
\newblock Latent video diffusion models for high-fidelity video generation with
  arbitrary lengths.
\newblock \emph{arXiv preprint arXiv:2211.13221}, 2022.

\bibitem[Ho et~al.(2020)Ho, Jain, and Abbeel]{ho2021denoising}
Jonathan Ho, Ajay Jain, and Pieter Abbeel.
\newblock Denoising diffusion probabilistic models.
\newblock In \emph{Advances in Neural Information Processing Systems}, 2020.

\bibitem[Ho et~al.(2022{\natexlab{a}})Ho, Chan, Saharia, Whang, Gao, Gritsenko,
  Kingma, Poole, Norouzi, Fleet, and Salimans]{ho2022imagen}
Jonathan Ho, William Chan, Chitwan Saharia, Jay Whang, Ruiqi Gao, Alexey
  Gritsenko, Diederik~P. Kingma, Ben Poole, Mohammad Norouzi, David~J. Fleet,
  and Tim Salimans.
\newblock Imagen video: High definition video generation with diffusion models.
\newblock \emph{arXiv preprint arXiv:2210.02303}, 2022{\natexlab{a}}.

\bibitem[Ho et~al.(2022{\natexlab{b}})Ho, Salimans, Gritsenko, Chan, Norouzi,
  and Fleet]{ho2022video}
Jonathan Ho, Tim Salimans, Alexey Gritsenko, William Chan, Mohammad Norouzi,
  and David~J Fleet.
\newblock Video diffusion models.
\newblock In \emph{Advances in Neural Information Processing Systems},
  2022{\natexlab{b}}.

\bibitem[Hong et~al.(2023)Hong, Ding, Zheng, Liu, and Tang]{hong2023cogvideo}
Wenyi Hong, Ming Ding, Wendi Zheng, Xinghan Liu, and Jie Tang.
\newblock Cogvideo: Large-scale pretraining for text-to-video generation via
  transformers.
\newblock In \emph{International Conference on Learning Representations}, 2023.

\bibitem[H{\"o}ppe et~al.(2022)H{\"o}ppe, Mehrjou, Bauer, Nielsen, and
  Dittadi]{hoppe2022diffusion}
Tobias H{\"o}ppe, Arash Mehrjou, Stefan Bauer, Didrik Nielsen, and Andrea
  Dittadi.
\newblock Diffusion models for video prediction and infilling.
\newblock \emph{Transactions on Machine Learning Research}, 2022.

\bibitem[Hsieh et~al.(2018)Hsieh, Liu, Huang, Fei-Fei, and
  Niebles]{hsieh2018learning}
Jun-Ting Hsieh, Bingbin Liu, De-An Huang, Li~Fei-Fei, and Juan~Carlos Niebles.
\newblock Learning to decompose and disentangle representations for video
  prediction.
\newblock In \emph{Advances in Neural Information Processing Systems}, 2018.

\bibitem[Huang \& Belongie(2017)Huang and Belongie]{huang2017arbitrary}
Xun Huang and Serge Belongie.
\newblock Arbitrary style transfer in real-time with adaptive instance
  normalization.
\newblock In \emph{IEEE International Conference on Computer Vision}, 2017.

\bibitem[Jiang et~al.(2023)Jiang, Yang, Koh, Wu, Loy, and
  Liu]{jiang2023text2performer}
Yuming Jiang, Shuai Yang, Tong~Liang Koh, Wayne Wu, Chen~Change Loy, and Ziwei
  Liu.
\newblock {Text2Performer}: Text-driven human video generation.
\newblock \emph{arXiv preprint arXiv:2303.13495}, 2023.

\bibitem[Kalchbrenner et~al.(2017)Kalchbrenner, Oord, Simonyan, Danihelka,
  Vinyals, Graves, and Kavukcuoglu]{kalchbrenner2017video}
Nal Kalchbrenner, A{\"a}ron Oord, Karen Simonyan, Ivo Danihelka, Oriol Vinyals,
  Alex Graves, and Koray Kavukcuoglu.
\newblock Video pixel networks.
\newblock In \emph{International Conference on Machine Learning}, 2017.

\bibitem[Karras et~al.(2020)Karras, Laine, Aittala, Hellsten, Lehtinen, and
  Aila]{karras2020analyzing}
Tero Karras, Samuli Laine, Miika Aittala, Janne Hellsten, Jaakko Lehtinen, and
  Timo Aila.
\newblock Analyzing and improving the image quality of {StyleGAN}.
\newblock In \emph{IEEE Conference on Computer Vision and Pattern Recognition},
  2020.

\bibitem[Karras et~al.(2022)Karras, Aittala, Aila, and Laine]{karras2022edm}
Tero Karras, Miika Aittala, Timo Aila, and Samuli Laine.
\newblock Elucidating the design space of diffusion-based generative models.
\newblock In \emph{Advances in Neural Information Processing Systems}, 2022.

\bibitem[Khachatryan et~al.(2023)Khachatryan, Movsisyan, Tadevosyan, Henschel,
  Wang, Navasardyan, and Shi]{text2video-zero}
Levon Khachatryan, Andranik Movsisyan, Vahram Tadevosyan, Roberto Henschel,
  Zhangyang Wang, Shant Navasardyan, and Humphrey Shi.
\newblock Text2video-zero: Text-to-image diffusion models are zero-shot video
  generators.
\newblock \emph{arXiv preprint arXiv:2303.13439}, 2023.

\bibitem[Kim et~al.(2022)Kim, Yu, Lee, and Shin]{kim2022scalable}
Subin Kim, Sihyun Yu, Jaeho Lee, and Jinwoo Shin.
\newblock Scalable neural video representations with learnable positional
  features.
\newblock In \emph{Advances in Neural Information Processing Systems}, 2022.

\bibitem[Kim et~al.(2023)Kim, Lee, Choi, Jeong, Sohn, and
  Shin]{kim2023collaborative}
Subin Kim, Kyungmin Lee, June~Suk Choi, Jongheon Jeong, Kihyuk Sohn, and Jinwoo
  Shin.
\newblock Collaborative score distillation for consistent visual synthesis.
\newblock In \emph{Advances in Neural Information Processing Systems}, 2023.

\bibitem[Kingma \& Ba(2015)Kingma and Ba]{kingma2014adam}
Diederik~P Kingma and Jimmy Ba.
\newblock Adam: A method for stochastic optimization.
\newblock In \emph{International Conference on Learning Representations}, 2015.

\bibitem[Kumar et~al.(2020)Kumar, Babaeizadeh, Erhan, Finn, Levine, Dinh, and
  Kingma]{kumar2020videoflow}
Manoj Kumar, Mohammad Babaeizadeh, Dumitru Erhan, Chelsea Finn, Sergey Levine,
  Laurent Dinh, and Durk Kingma.
\newblock Videoflow: A conditional flow-based model for stochastic video
  generation.
\newblock In \emph{International Conference on Learning Representations}, 2020.

\bibitem[Lee et~al.(2018)Lee, Zhang, Ebert, Abbeel, Finn, and
  Levine]{lee2018stochastic}
Alex~X Lee, Richard Zhang, Frederik Ebert, Pieter Abbeel, Chelsea Finn, and
  Sergey Levine.
\newblock Stochastic adversarial video prediction.
\newblock \emph{arXiv preprint arXiv:1804.01523}, 2018.

\bibitem[Lee et~al.(2021)Lee, Jung, Zhang, Chen, Koh, Huang, Yoon, Lee, and
  Hong]{lee2021revisiting}
Wonkwang Lee, Whie Jung, Han Zhang, Ting Chen, Jing~Yu Koh, Thomas Huang,
  Hyungsuk Yoon, Honglak Lee, and Seunghoon Hong.
\newblock Revisiting hierarchical approach for persistent long-term video
  prediction.
\newblock In \emph{International Conference on Learning Representations}, 2021.

\bibitem[Li et~al.(2023{\natexlab{a}})Li, Chu, Wu, Yuan, Liu, Zhang, Li, Feng,
  Ding, and Wang]{li2023videogen}
Xin Li, Wenqing Chu, Ye~Wu, Weihang Yuan, Fanglong Liu, Qi~Zhang, Fu~Li,
  Haocheng Feng, Errui Ding, and Jingdong Wang.
\newblock {VideoGen}: A reference-guided latent diffusion approach for high
  definition text-to-video generation.
\newblock \emph{arXiv preprint arXiv:2309.00398}, 2023{\natexlab{a}}.

\bibitem[Li et~al.(2023{\natexlab{b}})Li, Tucker, Snavely, and
  Holynski]{li2023generative}
Zhengqi Li, Richard Tucker, Noah Snavely, and Aleksander Holynski.
\newblock Generative image dynamics.
\newblock \emph{arXiv preprint arXiv:2309.07906}, 2023{\natexlab{b}}.

\bibitem[Lin et~al.(2023)Lin, Zala, Cho, and Bansal]{lin2023videodirectorgpt}
Han Lin, Abhay Zala, Jaemin Cho, and Mohit Bansal.
\newblock Videodirectorgpt: Consistent multi-scene video generation via
  llm-guided planning.
\newblock \emph{arXiv preprint arXiv:2309.15091}, 2023.

\bibitem[Lu et~al.(2023)Lu, Yang, Fei, Huo, Lu, Luo, and Ding]{lu2023vdt}
Haoyu Lu, Guoxing Yang, Nanyi Fei, Yuqi Huo, Zhiwu Lu, Ping Luo, and Mingyu
  Ding.
\newblock Vdt: An empirical study on video diffusion with transformers.
\newblock \emph{arXiv preprint arXiv:2305.13311}, 2023.

\bibitem[Luc et~al.(2020)Luc, Clark, Dieleman, Casas, Doron, Cassirer, and
  Simonyan]{luc2020transformation}
Pauline Luc, Aidan Clark, Sander Dieleman, Diego de~Las Casas, Yotam Doron,
  Albin Cassirer, and Karen Simonyan.
\newblock Transformation-based adversarial video prediction on large-scale
  data.
\newblock \emph{arXiv preprint arXiv:2003.04035}, 2020.

\bibitem[Luo et~al.(2023)Luo, Chen, Zhang, Huang, Wang, Shen, Zhao, Zhou, and
  Tan]{luo2023videofusion}
Zhengxiong Luo, Dayou Chen, Yingya Zhang, Yan Huang, Liang Wang, Yujun Shen,
  Deli Zhao, Jingren Zhou, and Tieniu Tan.
\newblock Videofusion: Decomposed diffusion models for high-quality video
  generation.
\newblock In \emph{IEEE Conference on Computer Vision and Pattern Recognition},
  2023.

\bibitem[Meng et~al.(2022)Meng, He, Song, Song, Wu, Zhu, and
  Ermon]{meng2022sdedit}
Chenlin Meng, Yutong He, Yang Song, Jiaming Song, Jiajun Wu, Jun-Yan Zhu, and
  Stefano Ermon.
\newblock {SDEdit}: Guided image synthesis and editing with stochastic
  differential equations.
\newblock In \emph{International Conference on Learning Representations}, 2022.

\bibitem[Miech et~al.(2019)Miech, Zhukov, Alayrac, Tapaswi, Laptev, and
  Sivic]{miech2019howto100m}
Antoine Miech, Dimitri Zhukov, Jean-Baptiste Alayrac, Makarand Tapaswi, Ivan
  Laptev, and Josef Sivic.
\newblock Howto100m: Learning a text-video embedding by watching hundred
  million narrated video clips.
\newblock In \emph{IEEE International Conference on Computer Vision}, 2019.

\bibitem[Molad et~al.(2023)Molad, Horwitz, Valevski, Acha, Matias, Pritch,
  Leviathan, and Hoshen]{molad2023dreamix}
Eyal Molad, Eliahu Horwitz, Dani Valevski, Alex~Rav Acha, Yossi Matias, Yael
  Pritch, Yaniv Leviathan, and Yedid Hoshen.
\newblock Dreamix: Video diffusion models are general video editors.
\newblock \emph{arXiv preprint arXiv:2302.01329}, 2023.

\bibitem[Munoz et~al.(2021)Munoz, Zolfaghari, Argus, and
  Brox]{munoz2021temporal}
Andres Munoz, Mohammadreza Zolfaghari, Max Argus, and Thomas Brox.
\newblock Temporal shift {GAN} for large scale video generation.
\newblock In \emph{IEEE/CVF Winter Conference on Applications of Computer
  Vision}, 2021.

\bibitem[Ni et~al.(2023)Ni, Shi, Li, Huang, and Min]{ni2023conditional}
Haomiao Ni, Changhao Shi, Kai Li, Sharon~X Huang, and Martin~Renqiang Min.
\newblock Conditional image-to-video generation with latent flow diffusion
  models.
\newblock In \emph{IEEE Conference on Computer Vision and Pattern Recognition},
  2023.

\bibitem[Nichol \& Dhariwal(2021)Nichol and Dhariwal]{nichol2021improved}
Alexander~Quinn Nichol and Prafulla Dhariwal.
\newblock Improved denoising diffusion probabilistic models.
\newblock In \emph{International Conference on Machine Learning}, 2021.

\bibitem[Peebles \& Xie(2023)Peebles and Xie]{Peebles2022DiT}
William Peebles and Saining Xie.
\newblock Scalable diffusion models with transformers.
\newblock In \emph{IEEE International Conference on Computer Vision}, 2023.

\bibitem[Radford et~al.(2021)Radford, Kim, Hallacy, Ramesh, Goh, Agarwal,
  Sastry, Askell, Mishkin, Clark, Krueger, and Sutskever]{radford2021learning}
Alec Radford, Jong~Wook Kim, Chris Hallacy, Aditya Ramesh, Gabriel Goh,
  Sandhini Agarwal, Girish Sastry, Amanda Askell, Pamela Mishkin, Jack Clark,
  Gretchen Krueger, and Ilya Sutskever.
\newblock Learning transferable visual models from natural language
  supervision.
\newblock In \emph{International Conference on Machine Learning}, 2021.

\bibitem[Rakhimov et~al.(2020)Rakhimov, Volkhonskiy, Artemov, Zorin, and
  Burnaev]{rakhimov2020latent}
Ruslan Rakhimov, Denis Volkhonskiy, Alexey Artemov, Denis Zorin, and Evgeny
  Burnaev.
\newblock Latent video transformer.
\newblock \emph{arXiv preprint arXiv:2006.10704}, 2020.

\bibitem[Rombach et~al.(2022)Rombach, Blattmann, Lorenz, Esser, and
  Ommer]{rombach2021highresolution}
Robin Rombach, Andreas Blattmann, Dominik Lorenz, Patrick Esser, and Bj\"orn
  Ommer.
\newblock High-resolution image synthesis with latent diffusion models.
\newblock In \emph{IEEE Conference on Computer Vision and Pattern Recognition},
  2022.

\bibitem[Ruiz et~al.(2023)Ruiz, Li, Jampani, Pritch, Rubinstein, and
  Aberman]{ruiz2023dreambooth}
Nataniel Ruiz, Yuanzhen Li, Varun Jampani, Yael Pritch, Michael Rubinstein, and
  Kfir Aberman.
\newblock {Dreambooth}: Fine tuning text-to-image diffusion models for
  subject-driven generation.
\newblock In \emph{IEEE Conference on Computer Vision and Pattern Recognition},
  2023.

\bibitem[Saharia et~al.(2022)Saharia, Chan, Saxena, Li, Whang, Denton,
  Ghasemipour, Gontijo~Lopes, Karagol~Ayan, Salimans,
  et~al.]{saharia2022photorealistic}
Chitwan Saharia, William Chan, Saurabh Saxena, Lala Li, Jay Whang, Emily~L
  Denton, Kamyar Ghasemipour, Raphael Gontijo~Lopes, Burcu Karagol~Ayan, Tim
  Salimans, et~al.
\newblock Photorealistic text-to-image diffusion models with deep language
  understanding.
\newblock In \emph{Advances in Neural Information Processing Systems}, 2022.

\bibitem[Schuhmann et~al.(2022)Schuhmann, Beaumont, Vencu, Gordon, Wightman,
  Cherti, Coombes, Katta, Mullis, Wortsman, et~al.]{schuhmann2022laion}
Christoph Schuhmann, Romain Beaumont, Richard Vencu, Cade Gordon, Ross
  Wightman, Mehdi Cherti, Theo Coombes, Aarush Katta, Clayton Mullis, Mitchell
  Wortsman, et~al.
\newblock {LAION-5B}: An open large-scale dataset for training next generation
  image-text models.
\newblock In \emph{Advances in Neural Information Processing Systems}, 2022.

\bibitem[Seo et~al.(2022)Seo, Lee, Liu, James, and
  Abbeel]{seo2022autoregressive}
Younggyo Seo, Kimin Lee, Fangchen Liu, Stephen James, and Pieter Abbeel.
\newblock Autoregressive latent video prediction with high-fidelity image
  generator.
\newblock In \emph{IEEE International Conference on Image Processing}, 2022.

\bibitem[Singer et~al.(2023)Singer, Polyak, Hayes, Yin, An, Zhang, Hu, Yang,
  Ashual, Gafni, et~al.]{singer2022make}
Uriel Singer, Adam Polyak, Thomas Hayes, Xi~Yin, Jie An, Songyang Zhang, Qiyuan
  Hu, Harry Yang, Oron Ashual, Oran Gafni, et~al.
\newblock Make-a-video: Text-to-video generation without text-video data.
\newblock In \emph{International Conference on Learning Representations}, 2023.

\bibitem[Skorokhodov et~al.(2022)Skorokhodov, Tulyakov, and
  Elhoseiny]{skorokhodov2021stylegan}
Ivan Skorokhodov, Sergey Tulyakov, and Mohamed Elhoseiny.
\newblock {StyleGAN-V}: A continuous video generator with the price, image
  quality and perks of {StyleGAN2}.
\newblock In \emph{IEEE Conference on Computer Vision and Pattern Recognition},
  2022.

\bibitem[Song et~al.(2021{\natexlab{a}})Song, Meng, and
  Ermon]{song2021denoising}
Jiaming Song, Chenlin Meng, and Stefano Ermon.
\newblock Denoising diffusion implicit models.
\newblock In \emph{International Conference on Learning Representations},
  2021{\natexlab{a}}.

\bibitem[Song et~al.(2021{\natexlab{b}})Song, Sohl-Dickstein, Kingma, Kumar,
  Ermon, and Poole]{song2021scorebased}
Yang Song, Jascha Sohl-Dickstein, Diederik~P Kingma, Abhishek Kumar, Stefano
  Ermon, and Ben Poole.
\newblock Score-based generative modeling through stochastic differential
  equations.
\newblock In \emph{International Conference on Learning Representations},
  2021{\natexlab{b}}.

\bibitem[Soomro et~al.(2012)Soomro, Zamir, and Shah]{soomro2012ucf101}
Khurram Soomro, Amir~Roshan Zamir, and Mubarak Shah.
\newblock {UCF101}: A dataset of 101 human actions classes from videos in the
  wild.
\newblock \emph{arXiv preprint arXiv:1212.0402}, 2012.

\bibitem[Srivastava et~al.(2015)Srivastava, Mansimov, and
  Salakhudinov]{srivastava2015unsupervised}
Nitish Srivastava, Elman Mansimov, and Ruslan Salakhudinov.
\newblock Unsupervised learning of video representations using {LSTM}s.
\newblock In \emph{International Conference on Machine Learning}, 2015.

\bibitem[Tian et~al.(2021)Tian, Ren, Chai, Olszewski, Peng, Metaxas, and
  Tulyakov]{tian2021good}
Yu~Tian, Jian Ren, Menglei Chai, Kyle Olszewski, Xi~Peng, Dimitris~N Metaxas,
  and Sergey Tulyakov.
\newblock A good image generator is what you need for high-resolution video
  synthesis.
\newblock In \emph{International Conference on Learning Representations}, 2021.

\bibitem[Tulyakov et~al.(2018)Tulyakov, Liu, Yang, and
  Kautz]{tulyakov2018mocogan}
Sergey Tulyakov, Ming-Yu Liu, Xiaodong Yang, and Jan Kautz.
\newblock {MoCoGAN}: Decomposing motion and content for video generation.
\newblock In \emph{IEEE Conference on Computer Vision and Pattern Recognition},
  2018.

\bibitem[Unterthiner et~al.(2018)Unterthiner, van Steenkiste, Kurach, Marinier,
  Michalski, and Gelly]{unterthiner2018towards}
Thomas Unterthiner, Sjoerd van Steenkiste, Karol Kurach, Raphael Marinier,
  Marcin Michalski, and Sylvain Gelly.
\newblock Towards accurate generative models of video: A new metric \&
  challenges.
\newblock \emph{arXiv preprint arXiv:1812.01717}, 2018.

\bibitem[van~den Oord et~al.(2017)van~den Oord, Vinyals, and
  Kavukcuoglu]{van2017neural}
Aaron van~den Oord, Oriol Vinyals, and Koray Kavukcuoglu.
\newblock Neural discrete representation learning.
\newblock In \emph{Advances in Neural Information Processing Systems}, 2017.

\bibitem[Villegas et~al.(2017)Villegas, Yang, Hong, Lin, and
  Lee]{villegas2017decomposing}
Ruben Villegas, Jimei Yang, Seunghoon Hong, Xunyu Lin, and Honglak Lee.
\newblock Decomposing motion and content for natural video sequence prediction.
\newblock In \emph{International Conference on Learning Representations}, 2017.

\bibitem[Villegas et~al.(2019)Villegas, Pathak, Kannan, Erhan, Le, and
  Lee]{villegas2019high}
Ruben Villegas, Arkanath Pathak, Harini Kannan, Dumitru Erhan, Quoc~V Le, and
  Honglak Lee.
\newblock High fidelity video prediction with large stochastic recurrent neural
  networks.
\newblock In \emph{Advances in Neural Information Processing Systems}, 2019.

\bibitem[Villegas et~al.(2023)Villegas, Babaeizadeh, Kindermans, Moraldo,
  Zhang, Saffar, Castro, Kunze, and Erhan]{villegas2023phenaki}
Ruben Villegas, Mohammad Babaeizadeh, Pieter-Jan Kindermans, Hernan Moraldo,
  Han Zhang, Mohammad~Taghi Saffar, Santiago Castro, Julius Kunze, and Dumitru
  Erhan.
\newblock Phenaki: Variable length video generation from open domain textual
  descriptions.
\newblock In \emph{International Conference on Learning Representations}, 2023.

\bibitem[Wang et~al.(2023{\natexlab{a}})Wang, Yuan, Chen, Zhang, Wang, and
  Zhang]{wang2023modelscope}
Jiuniu Wang, Hangjie Yuan, Dayou Chen, Yingya Zhang, Xiang Wang, and Shiwei
  Zhang.
\newblock Modelscope text-to-video technical report.
\newblock \emph{arXiv preprint arXiv:2308.06571}, 2023{\natexlab{a}}.

\bibitem[Wang et~al.(2023{\natexlab{b}})Wang, Yang, Tuo, He, Zhu, Fu, and
  Liu]{wang2023videofactory}
Wenjing Wang, Huan Yang, Zixi Tuo, Huiguo He, Junchen Zhu, Jianlong Fu, and
  Jiaying Liu.
\newblock Videofactory: Swap attention in spatiotemporal diffusions for
  text-to-video generation.
\newblock \emph{arXiv preprint arXiv:2305.10874}, 2023{\natexlab{b}}.

\bibitem[Wang et~al.(2023{\natexlab{c}})Wang, Chen, Ma, Zhou, Huang, Wang,
  Yang, He, Yu, Yang, Guo, Wu, Si, Jiang, Chen, Loy, Dai, Lin, Qiao, and
  Liu]{wang2023lavie}
Yaohui Wang, Xinyuan Chen, Xin Ma, Shangchen Zhou, Ziqi Huang, Yi~Wang, Ceyuan
  Yang, Yinan He, Jiashuo Yu, Peiqing Yang, Yuwei Guo, Tianxing Wu, Chenyang
  Si, Yuming Jiang, Cunjian Chen, Chen~Change Loy, Bo~Dai, Dahua Lin, Yu~Qiao,
  and Ziwei Liu.
\newblock {LAVIE}: High-quality video generation with cascaded latent diffusion
  models.
\newblock \emph{arXiv preprint arXiv:2309.15103}, 2023{\natexlab{c}}.

\bibitem[Weissenborn et~al.(2020)Weissenborn, T{\"a}ckstr{\"o}m, and
  Uszkoreit]{weissenborn2020scaling}
Dirk Weissenborn, Oscar T{\"a}ckstr{\"o}m, and Jakob Uszkoreit.
\newblock Scaling autoregressive video models.
\newblock In \emph{International Conference on Learning Representations}, 2020.

\bibitem[Wu et~al.(2021)Wu, Huang, Zhang, Li, Ji, Yang, Sapiro, and
  Duan]{wu2021godiva}
Chenfei Wu, Lun Huang, Qianxi Zhang, Binyang Li, Lei Ji, Fan Yang, Guillermo
  Sapiro, and Nan Duan.
\newblock {GODIVA}: Generating open-domain videos from natural descriptions.
\newblock \emph{arXiv preprint arXiv:2104.14806}, 2021.

\bibitem[Wu et~al.(2022)Wu, Liang, Ji, Yang, Fang, Jiang, and Duan]{wu2022nuwa}
Chenfei Wu, Jian Liang, Lei Ji, Fan Yang, Yuejian Fang, Daxin Jiang, and Nan
  Duan.
\newblock N{\"u}wa: Visual synthesis pre-training for neural visual world
  creation.
\newblock In \emph{European Conference on Computer Vision}, 2022.

\bibitem[Xu et~al.(2016)Xu, Mei, Yao, and Rui]{xu2016msr}
Jun Xu, Tao Mei, Ting Yao, and Yong Rui.
\newblock Msr-vtt: A large video description dataset for bridging video and
  language.
\newblock In \emph{IEEE Conference on Computer Vision and Pattern Recognition},
  2016.

\bibitem[Xu et~al.(2023)Xu, Powers, Dror, Ermon, and Leskovec]{xu2023geometric}
Minkai Xu, Alexander~S Powers, Ron~O Dror, Stefano Ermon, and Jure Leskovec.
\newblock Geometric latent diffusion models for 3d molecule generation.
\newblock In \emph{International Conference on Machine Learning}, 2023.

\bibitem[Yan et~al.(2021)Yan, Zhang, Abbeel, and Srinivas]{yan2021videogpt}
Wilson Yan, Yunzhi Zhang, Pieter Abbeel, and Aravind Srinivas.
\newblock {VideoGPT}: Video generation using {VQ-VAE} and transformers.
\newblock \emph{arXiv preprint arXiv:2104.10157}, 2021.

\bibitem[Yang et~al.(2022)Yang, Srivastava, and Mandt]{yang2022diffusion}
Ruihan Yang, Prakhar Srivastava, and Stephan Mandt.
\newblock Diffusion probabilistic modeling for video generation.
\newblock \emph{arXiv preprint arXiv:2203.09481}, 2022.

\bibitem[Yu et~al.(2023{\natexlab{a}})Yu, Cheng, Sohn, Lezama, Zhang, Chang,
  Hauptmann, Yang, Hao, Essa, et~al.]{yu2023magvit}
Lijun Yu, Yong Cheng, Kihyuk Sohn, Jos{\'e} Lezama, Han Zhang, Huiwen Chang,
  Alexander~G Hauptmann, Ming-Hsuan Yang, Yuan Hao, Irfan Essa, et~al.
\newblock Magvit: Masked generative video transformer.
\newblock In \emph{IEEE Conference on Computer Vision and Pattern Recognition},
  2023{\natexlab{a}}.

\bibitem[Yu et~al.(2022)Yu, Tack, Mo, Kim, Kim, Ha, and Shin]{yu2022digan}
Sihyun Yu, Jihoon Tack, Sangwoo Mo, Hyunsu Kim, Junho Kim, Jung-Woo Ha, and
  Jinwoo Shin.
\newblock Generating videos with dynamics-aware implicit generative adversarial
  networks.
\newblock In \emph{International Conference on Learning Representations}, 2022.

\bibitem[Yu et~al.(2023{\natexlab{b}})Yu, Sohn, Kim, and Shin]{yu2023video}
Sihyun Yu, Kihyuk Sohn, Subin Kim, and Jinwoo Shin.
\newblock Video probabilistic diffusion models in projected latent space.
\newblock In \emph{IEEE Conference on Computer Vision and Pattern Recognition},
  2023{\natexlab{b}}.

\bibitem[Zeng et~al.(2022)Zeng, Vahdat, Williams, Gojcic, Litany, Fidler, and
  Kreis]{zeng2022lion}
Xiaohui Zeng, Arash Vahdat, Francis Williams, Zan Gojcic, Or~Litany, Sanja
  Fidler, and Karsten Kreis.
\newblock {LION}: Latent point diffusion models for 3d shape generation.
\newblock In \emph{Advances in Neural Information Processing Systems}, 2022.

\bibitem[Zhang et~al.(2023)Zhang, Wu, Liu, Zhao, Ran, Gu, Gao, and
  Shou]{zhang2023show1}
David~Junhao Zhang, Jay~Zhangjie Wu, Jia-Wei Liu, Rui Zhao, Lingmin Ran, Yuchao
  Gu, Difei Gao, and Mike~Zheng Shou.
\newblock Show-1: Marrying pixel and latent diffusion models for text-to-video
  generation.
\newblock \emph{arXiv preprint arXiv:2309.15818}, 2023.

\bibitem[Zhou et~al.(2022)Zhou, Wang, Yan, Lv, Zhu, and
  Feng]{zhou2022magicvideo}
Daquan Zhou, Weimin Wang, Hanshu Yan, Weiwei Lv, Yizhe Zhu, and Jiashi Feng.
\newblock Magicvideo: Efficient video generation with latent diffusion models.
\newblock \emph{arXiv preprint arXiv:2211.11018}, 2022.

\end{thebibliography}
\bibliographystyle{iclr2024_conference}

\appendix
\newpage
\section{More Discussion on Related Work}
\label{appen:related}
There are several recent works that have some similarities to \sname. In what follows, we discuss the relevant work and the differences with our method in detail.

\textbf{Motion-content decomposition.}
\sname is similar to many previous video GANs that generate videos via motion-content decomposition ~\citep{villegas2017decomposing,hsieh2018learning,tulyakov2018mocogan,tian2021good,munoz2021temporal,yu2022digan,skorokhodov2021stylegan}. To achieve generation with motion-content controllability, they sample random content and motion vectors and use them to generate overall style and underlying motions. We take a similar approach using pretrained image diffusion models by training a video autoencoder that decomposes video as a content frame (similar to content vectors) and motion latent representation. Here, note that \sname does not strictly decompose motion and contents in a given video, but the scheme is similar at a high level. {Moreover, \sname shares a similarity to several diffusion models for video generation~\citep{guo2023animatediff, jiang2023text2performer,ni2023conditional}. However, they have not achieved (a) an efficient generative modeling and (b) exploitation of pretrained image diffusion models simultaneously. Specifically, \citep{guo2023animatediff} exploits pretrained image diffusion models, but it deals with videos as cubic tensors, \citep{jiang2023text2performer,ni2023conditional} do not use pretrained image diffusion models, and \citep{ni2023conditional} uses flow for motion encoding, which is a high-dimensional cubic tensor as well. } 

\textbf{Video prediction.}
Our work also has a relationship between video prediction models that forecast future video frames given at previous frames as input \citep{srivastava2015unsupervised,finn2016unsupervised,denton2017unsupervised,babaeizadeh2018stochastic,denton2018stochastic,lee2018stochastic,villegas2019high,kumar2020videoflow,franceschi2020stochastic,luc2020transformation,lee2021revisiting,seo2022autoregressive}. Similar to our method, some video prediction methods predict video frames as low-dimensional latent space rather than raw pixel space to increase the window size and achieve efficiency. The main difference between our method and these approaches is in the input condition: our method provides a content frame as an input, whereas they provide initial frames as an input.

\textbf{Difference with PVDM.}
Our model shares a similarity to recent latent diffusion models for videos~\citep{he2022lvdm,zhou2022magicvideo,yu2023video}. In particular, our method is quite similar to PVDM~\citep{yu2023video}; this work also proposes a latent video diffusion model to target unconditional video generation based on proposing a video encoding scheme to decompose them as triplane latents. In this work, each latent in triplane latents lies in a different space from the video frames. Thus, it is difficult to exploit pretrained image models. Different from this work, our primary focus is on conditional video generation, and we introduce the ``content frames'' concept to exploit pretrained image diffusion while avoiding handling giant cubic video tensors. 

\textbf{Difference with VideoFusion.}
Our work shares a similarity to VideoFusion~\citep{luo2023videofusion}. Unlike conventional approaches that add temporal layers to the image diffusion models for achieving T2V generation, VideoFusion also considers the training of an additional diffusion model in addition to the pretrained image model for generating videos. However, in contrast to \sname, their primary focus is not on achieving dimension reduction for improving computation and memory efficiency.

\textbf{Discussion with concurrent works.}
Show-1~\citep{zhang2023show1} also considers an efficient T2V generation via a mixture of pixel-level image diffusion models and latent image diffusion models. LAVIE~\citep{wang2023lavie} considers video generation using cascaded latent diffusion models. VideoDirectorGPT~\citep{lin2023videodirectorgpt} proposes to combine large language models to generate text prompts for longer video generation. \citet{li2023generative} considers controllable video generation conditioned on a given image and the motion direction. Text2Video-Zero~\citep{text2video-zero} considers zero-shot video generation from a pretrained T2I model without any video data. However, neither of them considers the temporal coherency of videos for improving efficiency.

\newpage
\section{Implementation Details}
\label{appen:implementation}

\subsection{Datasets}
\label{appen:datasets}
\textbf{UCF-101} \citep{soomro2012ucf101} is a human video dataset that includes 101 different types of human actions. Each video consists of frames with 320$\times$240 resolution with varying video lengths. The dataset contains 13,320 videos in total; 9,537 videos are in the train split, and the rest of them are in the test split. We only use train split for training and evaluation of the model, following the experimental setup used in recent video generation methods~\citep{yu2023magvit, singer2022make}. We resized all video frames to 64$\times$64 resolution frames and clipped them to a video length of 16. For the zero-shot evaluation, we resize each video to 512$\times$512 resolution.

\textbf{WebVid-10M} is a dataset that consists of 10,727,607 text-video pairs as training split. The dataset also contains a validation split that is composed of 5,000 text-video pairs. We use train split for training the model and use a validation set for evaluation. Since some videos in the training split are not available, we exclude these videos from training. We resize all video frames to 512$\times$1024 resolution and clip each video into a length of 16. 

\textbf{MSR-VTT} is a dataset consisting of 10,000 videos and corresponding captions. We use test split for zero-shot evaluation of our method, which contains 2,000 videos and corresponding text captions. We only use text captions to measure the alignment between text prompts and generated videos.

\subsection{Baselines}
\label{appen:baselines}
In what follows, we explain the main idea of baseline methods that we used for the evaluation.
\begin{itemize}[leftmargin=0.2in]
\item \textbf{DIGAN}~\citep{yu2022digan} presents a video GAN based on adapting the concept of implicit neural representations (or neural fields) into the generator.
\item \textbf{TATS}~\citep{ge2022long} presents a video autoencoder by extending an image autoencoder in VQGAN~\citep{esser2020taming} and trains an autoregressive Transformer for a generation.
\item \textbf{CogVideo}~\citep{hong2023cogvideo} presents a large-scale autoregressive Transformer for video generation extended from a pretrained model for images.
\item \textbf{Make-A-Video}~\citep{singer2022make} presents a method to achieve text-to-video generation without having text-video pairs but only with text-image pairs and video data. 
\item \textbf{VideoFusion}~\citep{luo2023videofusion} proposes a new extension scheme of pretrained image diffusion models for video generation by training an additional diffusion model that achieves frame-by-frame generation from intermediate noises from pretrained image diffusion models.
\item \textbf{MAGVIT}~\citep{yu2023video} proposes a non-autoregressive Transformer for videos, based on extending a non-autoregressive Transformer for images, MaskGiT~\citep{chang2022maskgit}.
\item \textbf{VideoFactory}~\citep{wang2023videofactory} proposes a new swapped cross-attention for better video diffusion models and introduces a large-scale text-video dataset.
\item \textbf{PYoCo}~\citep{ge2023preserve} explores noise prior to extend image diffusion models for better video generation, instead of starting from i.i.d. Gaussian noises. 
\item \textbf{LVDM}~\citep{he2022lvdm} extends latent image diffusion models by modeling video distribution in spatiotemporally downsampled latent space.
\item \textbf{ModelScope}~\citep{wang2023modelscope} trains a latent video diffusion model from a pretrained text-to-image diffusion model by adding several temporal layers.
\item \textbf{VideoLDM}~\citep{blattmann2023align} proposes an extension scheme from text-to-image diffusion models to text-to-video diffusion models by adding temporal layers in pretrained diffusion models and pretrained image autoencoder.
\item \textbf{VideoGen}~\citep{li2023videogen} generates video from a given image and text prompts by generating motion latent representation inspired by flow-based temporal upsampling.
\item \textbf{GOVIDA}~\citep{wu2021godiva} trains an autoregressive Transformer with large-scale text-video pair datasets, HowTo100M~\citep{miech2019howto100m}.
\item \textbf{N\"UWA}~\citep{wu2022nuwa} extends the approaches in GODIVA to multi-modality, including images, text, and videos.
\end{itemize}

\subsection{Training Details}
\label{appen:training}
\textbf{Autoencoder}.
In all experiments, we use the Adam optimizer~\citep{kingma2014adam} with a learning rate of 1e-5, $(\beta_1, \beta_2)=(0.5, 0.9)$ without weight decay. We use 8 NVIDIA A100 80GB GPUs for training with a batch size of 24, and it takes $\sim$1 week for the convergence {with 1024$\times$512 resolution videos}. For backbone networks, we use TimeSFormer~\citep{bertasius2021space} following PVDM~\citep{yu2023video}, and we use a single convolutional layer as pre- and post-layers at the end of the encoder and at the beginning of the decoder. We provide other hyperparameters related to the model configurations in Table~\ref{tab:ae_params}:
\begin{table}[ht!]
\centering\small
\caption{Hyperparameters related to our autoencoder. Hidden dim. (hidden dimension), depth, head dim. (head dimension), and num. head (number of heads) denote the channel, the depth, and the dimension and the number of heads used in the attention layer used in the backbone TimeSformer.}
 \resizebox{\textwidth}{!}{
\begin{tabular}{lccccccccccc}
    \toprule
      & $\bx$ channel & Height & Width & \multirow{2}{*}{{\begin{minipage}[t]{0.145\textwidth}{\centering Input patch size \\$(H/H', W/W')$}\end{minipage}}} & Hidden dim. & $\bz$ channel &  \\
     Dataset & {$C$} & {$H'$} & {$W'$} & & {$C'$} & {$D$} & {Depth} & {Head dim.} & {Num. head} \\
     \midrule
     UCF-101 & 4 & 32 & 32 & (2, 2) & {384} & {8} & {12} & {64} & {8}\\
     WebVid-10M (SD1.5) & 4 & 32 & 32 & (2, 2) & {384} & {8} & {12} & {64} & {8}\\
     WebVid-10M (SD2.1) & 4 & 32 & 64 & (2, 2) & {384} & {8} & {12} & {64} & {8}\\
     \bottomrule
\end{tabular}
}
\vspace{-0.15in}
\label{tab:ae_params}
\end{table}

\textbf{Motion diffusion model.}
We use the Adam optimizer with a learning rate of 1e-4, $(\beta_1, \beta_2)=(0.9, 0.999)$ and without weight decay. We use 8 and 32 NVIDIA A100 80GB GPUs to train the model on UCF-101 and WebVid (respectively) and use a batch size of 256. It takes 3-4 days for the convergence {with 1024$\times$512 resolution videos}. Since the dimension of $\bc$ of the text encoder and the hidden dimension in the motion diffusion model are different, we add and train a linear projection layer that maps the text hidden feature to the vector with the same dimension as the DiT hidden dimension. Our motion diffusion model implementation heavily follows the official implementation of DiT~\citep{Peebles2022DiT}, including hyperparameters and training objectives used.\footnote{\small{\url{https://github.com/facebookresearch/DiT}}} We provide other hyperparameters related to the model configurations in Table~\ref{tab:motion_params}.
\begin{table}[h!]
\vspace{-0.06in}
\centering\small
\caption{Hyperparameters related to our motion diffusion model.}
\vspace{-0.06in}
\resizebox{\textwidth}{!}{
\begin{tabular}{lccccccccccc}
    \toprule
    Dataset & Config. & Key. patch size & Input patch size & Text Encoder & Epochs & Ema\\
    \midrule
    UCF-101 & DiT-L/2 & 4 & 2 & - & 3000 & 0.999\\
    WebVid-10M (SD1.5) & DiT-XL/2 & 4 & 2 & OpenClip (ViT/H) & 3& 0.999\\
    WebVid-10M (SD2.1) & DiT-XL/2 & 8 & 2 & OpenClip (ViT/H) & 3& 0.999\\
     \bottomrule
\end{tabular}
}
\vspace{-0.15in}
\label{tab:motion_params}
\end{table}

\textbf{Content frame diffusion model.}
We use the Adam optimizer with a learning rate of 1e-4, $(\beta_1, \beta_2)=(0.9, 0.999)$ and without weight decay. We use 16 and 64 NVIDIA A100 80GB GPUs to train the model on UCF-101 and WebVid (respectively) and use a batch size of 256. For UCF-101 experiments, we train the model from scratch for a fair comparison with baselines. For WebVid-10M experiments, we fine-tune Stable Diffusion 1.5 and 2.1-base~\citep{rombach2021highresolution} for zero-shot evaluation on UCF-101 and other evaluations (respectively), following the recent text-to-video generation works that use pretrained image models~\citep{wang2023modelscope}. {It takes 3-4 days for the convergence with 1024$\times$512 resolution video frames due to high dimensionality of video frames.} We provide other hyperparameters related to the model configurations in Table~\ref{tab:keyframe_params}.
\begin{table}[ht!]
\centering\small
\vspace{-0.03in}
\caption{Hyperparameters related to our content frame diffusion model.}
\vspace{-0.06in}
\begin{tabular}{lccccccccccc}
    \toprule
    Dataset & Model & Epochs & Ema\\
    \midrule
    UCF-101 (non-zero-shot) & DiT-XL/2 & 3000 & 0.999\\

    WebVid-10M (T2V) & SD 2.1 (base) & 3 & 0.999\\
     \bottomrule
\end{tabular}
\label{tab:keyframe_params}
\vspace{-0.22in}
\end{table}

\subsection{Metrics}
\label{appen:metrics}
\textbf{Sampler.}
For both motion diffusion models and content frame diffusion models, we use the DDIM~\citep{song2021denoising} sampler. We use $\eta=0.0$ for both models (\ie, without additional random noises in sampling), and we use the number of steps as 100 and 50 for the motion diffusion model and the content frame diffusion model, respectively. For the content frame diffusion model, we use the classifier guidance scale $w=4.0$ on UCF-101 and $w=7.5$ on text-to-video generation.

\textbf{CLIPSIM.}
Following the protocol used in most text-to-video generation work~\citep{singer2022make,wang2023videofactory}, we calculate CLIP scores~\citep{wu2021godiva} between a text prompt and generated video frames and report the average between them. Specifically, we each video frame into an image of resolution 224$\times$224 and use it as an input to the CLIP image encoder. Following VideoLDM~\citep{blattmann2023align}, we use the ViT-B/32 CLIP model~\citep{radford2021learning}.

\begin{figure}[h!]
    \centering
    \includegraphics[width=\linewidth]
    {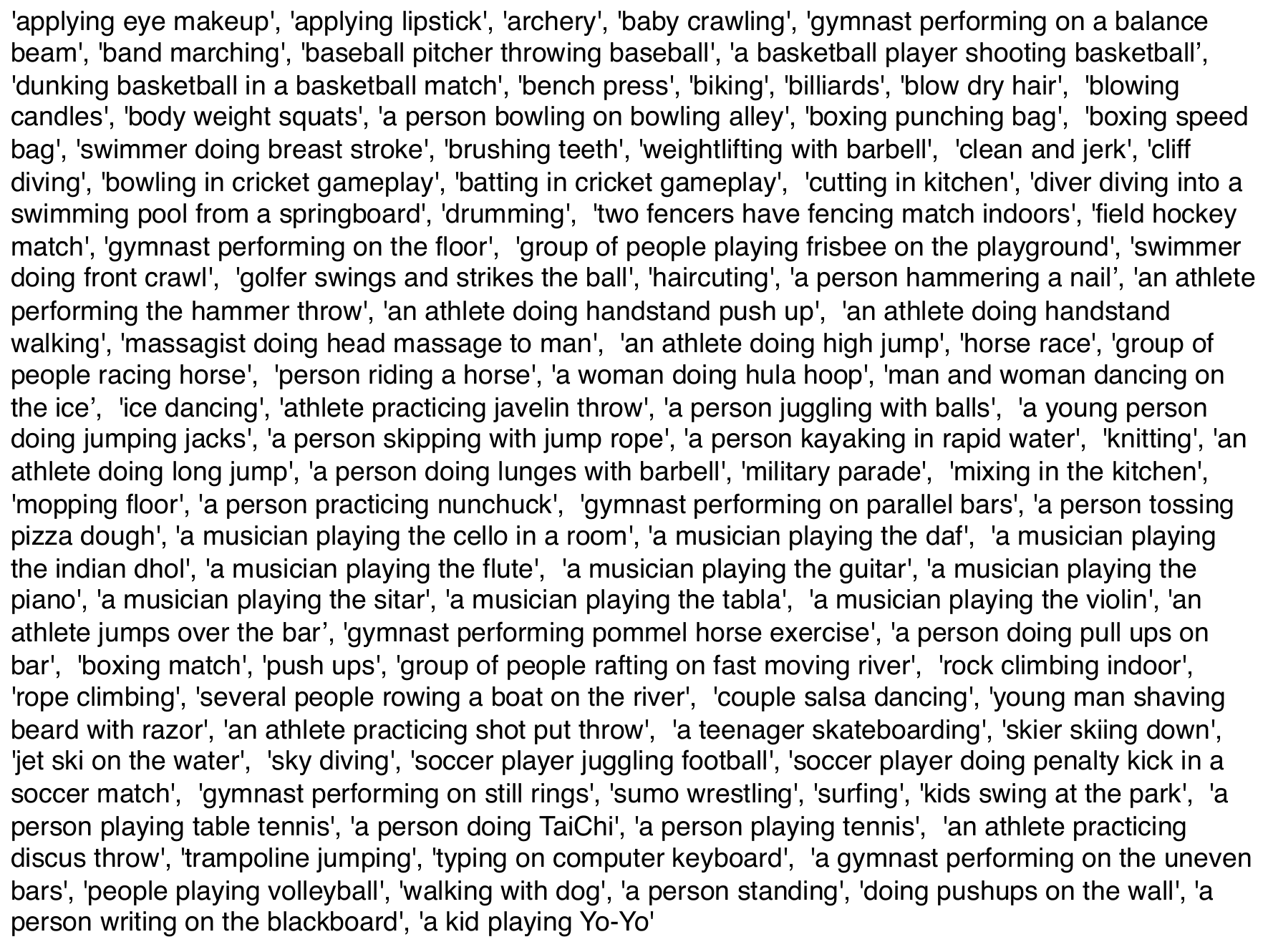}
    \vspace{-0.05in}
    \caption{Text prompts used for zero-shot evaluation on UCF-101.}
    \label{fig:prompts}
\end{figure}

\textbf{FVD.} 
For Fr\'echet video distance (FVD; \citealt{unterthiner2018towards}), we mainly follow the recently fixed evaluation protocol proposed by StyleGAN-V~\citep{skorokhodov2021stylegan}. Specifically, this protocol samples a single random video clip from each video and extracts the feature using a pretrained I3D network~\citep{carreira2017quo}. For UCF-101 (non-zero-shot), we consider representative scenarios for evaluation: 2,048 real/fake samples (used in most previous methods such as DIGAN~\citep{yu2022digan}) and 10,000 fake samples and clips in the training set (used in recent large-scale video generation methods such as MAGVIT~\citep{yu2023magvit}). For zero-shot evaluation on UCF-101, we use the common protocol to generate 100 videos per class and calculate the fake statistics~\citep{singer2022make}. For text prompts, we use the same text prompt used in PYoCo~\citep{ge2023preserve}, as shown in Figure~\ref{fig:prompts}. For non-zero-shot evaluation on WebVid-10M, we use 5,000 video clips and text prompts in the validation set to calculate real and fake statistics, respectively. For FPS, we follow the exact same setup in the concurrent VideoLDM~\citep{blattmann2023align}.

\newpage
\section{More qualitative results}
\vspace{-0.05in}
\label{appen:more_qual}
\begin{figure}[ht!]
    \centering
    \includegraphics[width=.95\linewidth]{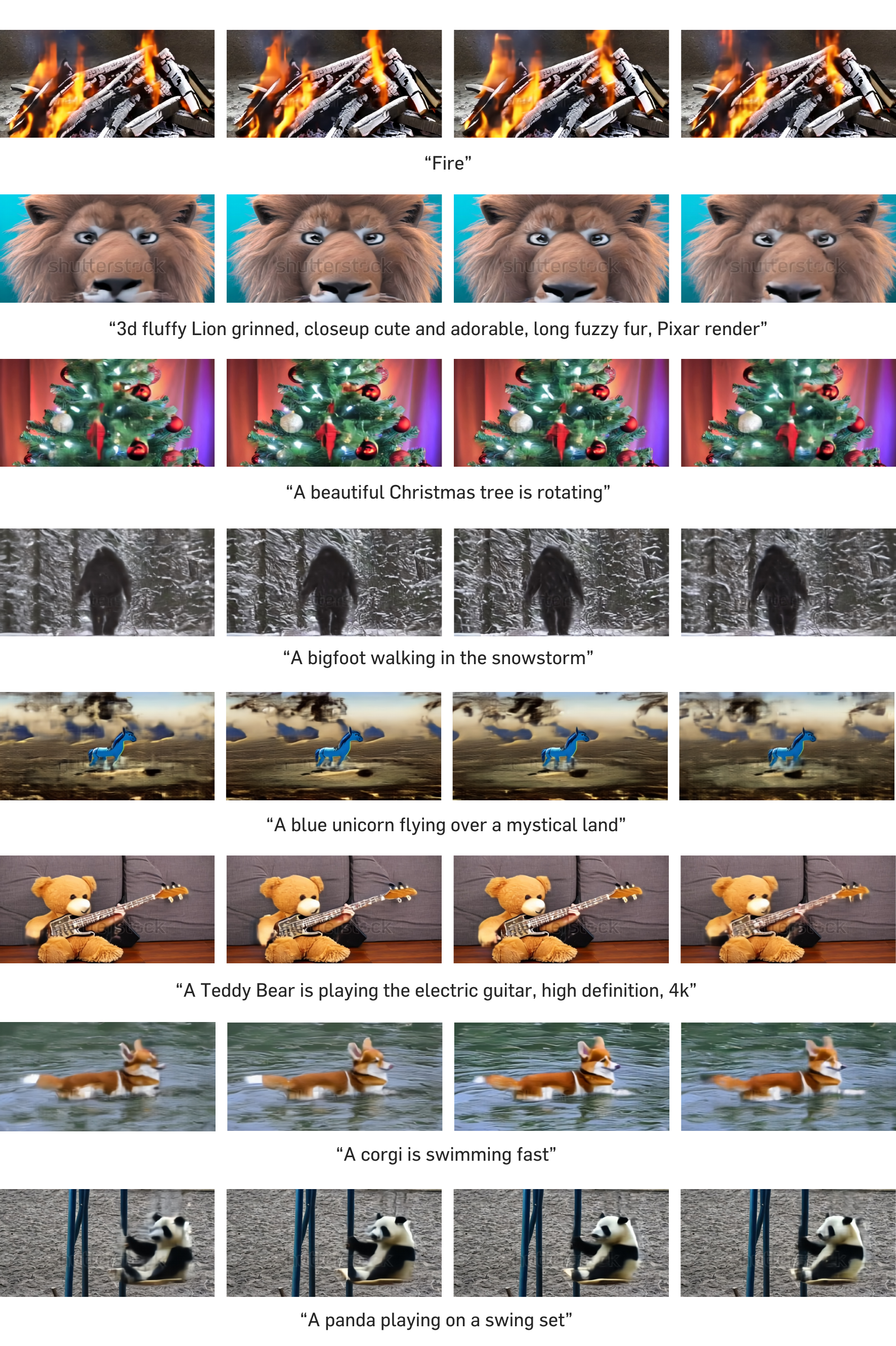}
    \caption{\textbf{512$\times$1024 resolution, 16-frame text-to-video generation results} from our \sname.}
\end{figure}
\begin{figure}[ht!]
    \centering
    \includegraphics[width=.95\linewidth]{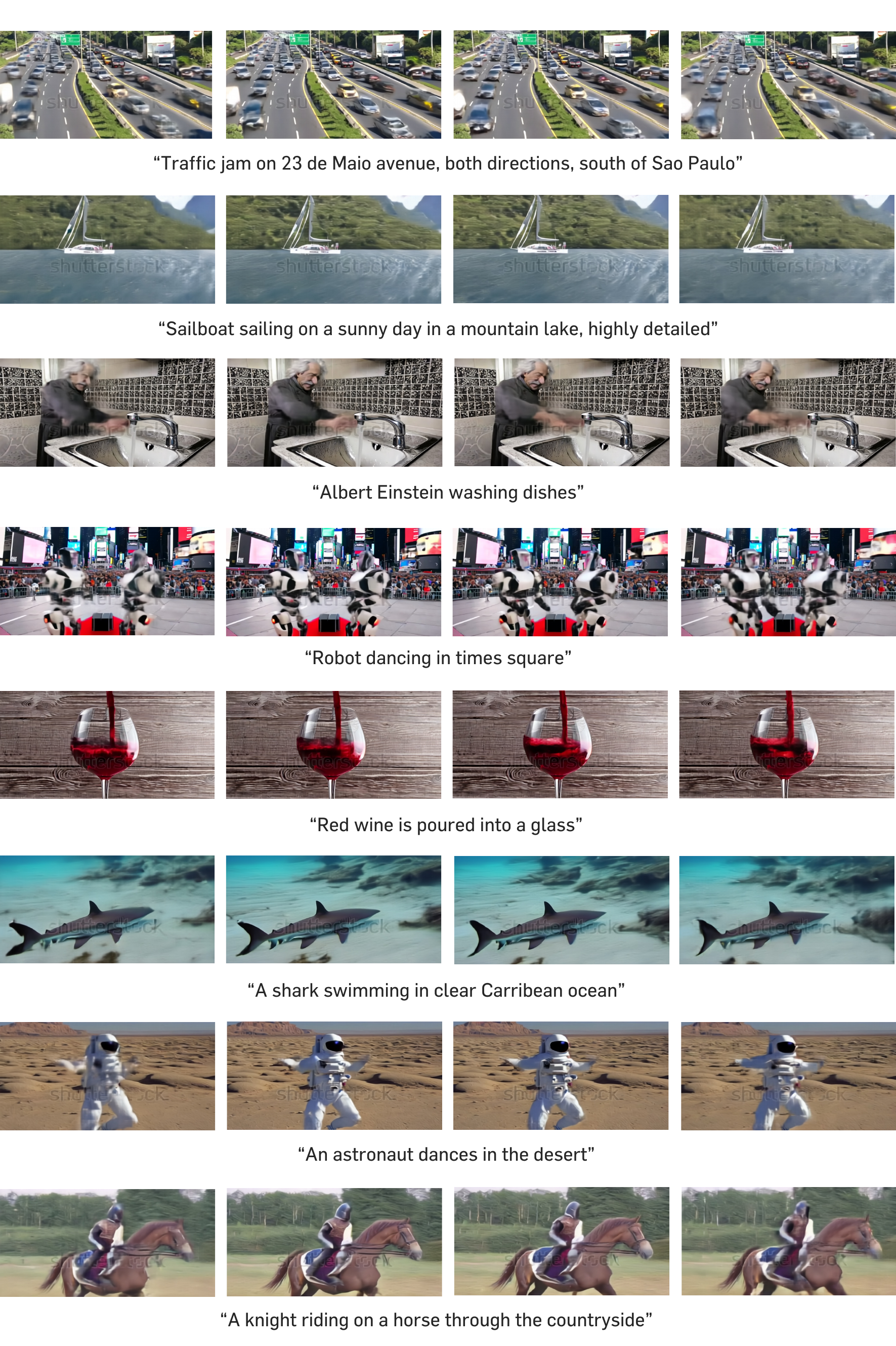}
    \caption{\textbf{512$\times$1024 resolution, 16-frame text-to-video generation results} from our \sname.}
\end{figure}
\begin{figure}[ht!]
    \centering
    \includegraphics[width=.95\linewidth]{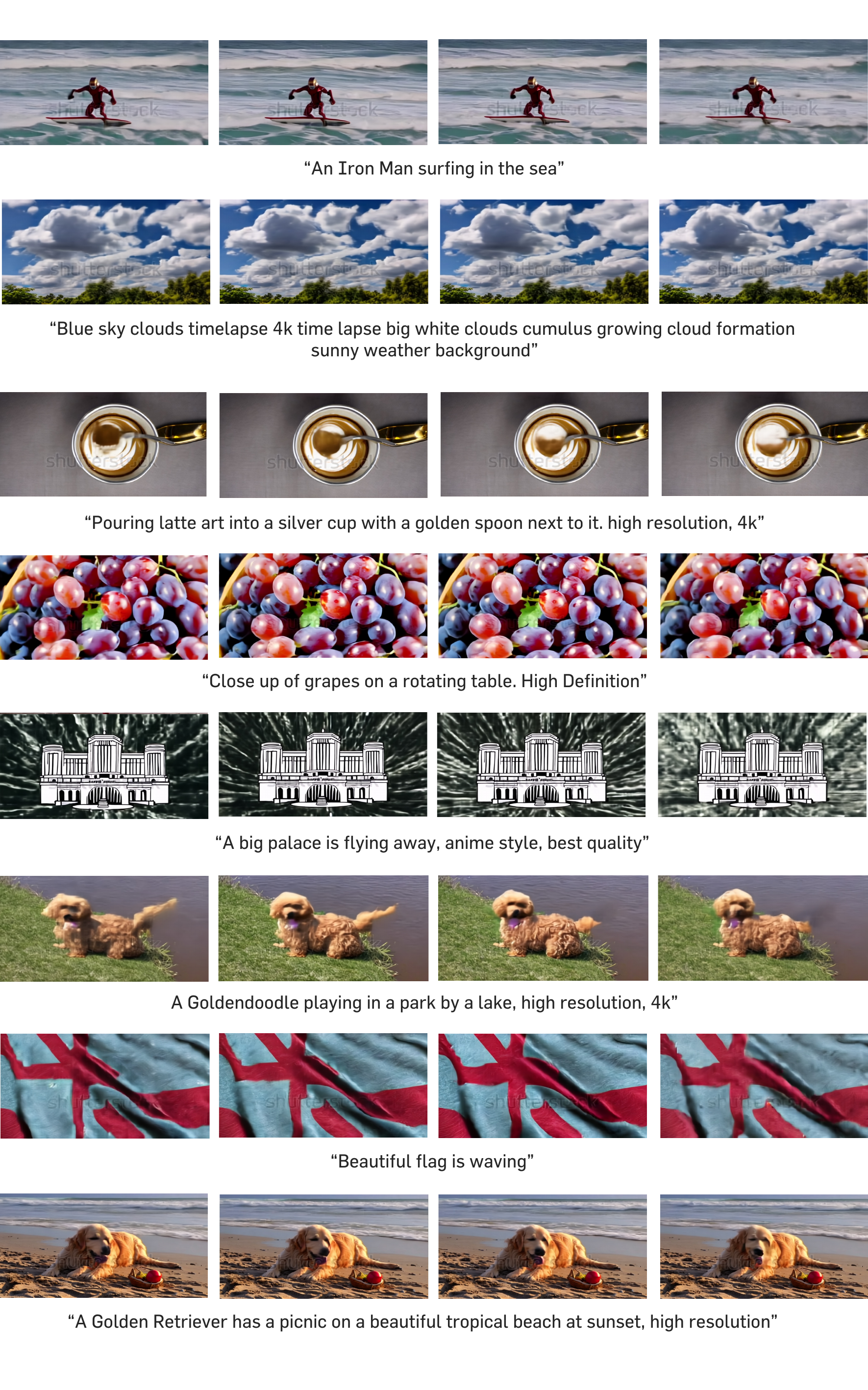}
    \vspace{-0.25in}
    \caption{\textbf{512$\times$1024 resolution, 16-frame text-to-video generation results} from our \sname.}
\end{figure}
\clearpage
\newpage
\section{More Details in Efficiency Analysis}
\label{appen:efficiency}
\textbf{Evaluation details.}
For a fair comparison with baselines, we mainly use the official implementation provided in LVDM~\citep{he2022lvdm}\footnote{\url{https://github.com/YingqingHe/LVDM}} and ModelScope~\citep{wang2023modelscope}.\footnote{\url{https://huggingface.co/spaces/damo-vilab/modelscope-text-to-video-synthesis}} For LVDM, we use \texttt{text2video.yaml} provided by the official implementation that uses a spatial downsample factor of 8 and a temporal downsample factor of 1. We adjust all factors and techniques that affect memory and computation dramatically. First, all values are measured with the same machine with a single NVIDIA A100 40GB/80GB GPU. Moreover, we use mixed precision operation and memory-efficient attention mechanisms in all baselines.\footnote{\url{https://github.com/facebookresearch/xformers}} We also use half precision (fp16) for measuring the efficiency in sampling. Finally, we disable gradient checkpointing for all baselines in measuring the training efficiency. For measuring floating point operations (FLOPs), we use the Fvcore library that measures the FLOPs of Pytorch models.\footnote{\url{https://github.com/facebookresearch/fvcore}}

\textbf{Exact values.} In Table~\ref{tab:component_efficiency} and \ref{tab:sampling_efficiency}, we provide exact values of each bar in Figure \ref{fig:component_efficiency} and \ref{fig:sampling_efficiency}.

\begin{table}[h!]
    \centering\small
    \caption{Exact values of component-wise efficiency analysis.}
    \begin{tabular}{llccc}
    \toprule
    {Method} & {Model} & {TFLOPs} & {sec/step} & {Memory (GB)} \\
    \midrule
    {ModelScope} & {Diffusion} & {9.41} & {0.920} & {69.6} \\
    {LVDM} & {Diffusion} & {6.28} & {0.456} & {44.2} \\
    \midrule
    {} & {Autoencoder} & {0.77} & {0.109} & {8.93} \\
    {\sname} & {Content Frame Diffusion} & {0.34} & {0.151} & {21.5} \\
    {} & {Motion Diffusion} & {0.14} & {0.100} & {18.7} \\
    \bottomrule
    \end{tabular}
    \label{tab:component_efficiency}
\end{table}

\begin{table}[h!]
    \centering\small
    \caption{Exact values of sampling efficiency analysis.}
    \begin{tabular}{lcccccc}
    \toprule
     {} & \multicolumn{2}{c}{TFLOPs} & \multicolumn{2}{c}{Time (s)} & \multicolumn{2}{c}{Memory (GB)} \\
     \cmidrule(lr){2-3} \cmidrule(lr){4-5} \cmidrule(lr){6-7}
     {Method} & {Step=50} & {Step=100} & {Step=50} & {Step=100} & {Batch=1} & {Batch=4} \\
    \midrule
    {ModelScope}  & {939.0} & {1877.9} & {35.1} & {70.3} & {8.50} & {21.3} \\
    {LVDM}  & {625.6} & {1251.2} & {24.2} & {48.1} & {8.38} & {18.6} \\
    {\sname}  & {\phantom{0}46.8} & {\phantom{00}92.1} &  {3.13} & {6.05} & {5.56} & {8.57} \\
    \bottomrule
    \end{tabular}
    \label{tab:sampling_efficiency}
\end{table}

\textbf{Model size comparison.}
In Table~\ref{tab:modelsize}, we summarize the number of model parameters of recent text-to-video generation models. Our model has a similar size to ModelScope~\citep{wang2023modelscope} and is much smaller than popular, well-performing text-to-video generation models, \eg, VideoLDM and Imagen-Video. In this respect, we strongly believe the generation quality of our \sname will be much stronger if one enlarges the overall model sizes by adjusting model configurations.

\begin{table}[h!]
    \centering\small
    \caption{Model size analysis.}
    \resizebox{\textwidth}{!}{
    \begin{tabular}{lcccccccc}
    \toprule
    {Method} & {ModelScope} & {LVDM} & {VideoLDM} & {Imagen-Video} & {PyoCo} & CogVideo & {\sname (Ours)} \\
    \midrule
    {\# params.} & {1.7B} & {0.96B} & {3.1B} & {11.6B} & {2.6B} & {9B} & {1.6B} \\ 
    \bottomrule
    \end{tabular}
    }
    \label{tab:modelsize}
\end{table}
\newpage
\section{Sampling procedure}
We summarize the sampling procedure of \sname in Algorithm~\ref{algo:1}.
\begin{algorithm}[h]
\begin{spacing}{1.05}
\caption{\lname (\sname)}\label{algo:1}
\begin{algorithmic}[1]
\State Sample random Gaussian noise $\bar{\bx}_T \sim \mathcal{N}(\mathbf{0}_{\bar{\bx}}, \mathbf{I}_{\bar{\bx}})$,  $\bz_T \sim \mathcal{N}(\mathbf{0}_{\bz}, \mathbf{I}_{\bz})$.
\For{$t=T$ to $1$}
\State $\bm{\epsilon}_t = (1+w) \bm{\epsilon}_{\bm{\theta}_I}(\bar{\bx}_t, \bc, t) 
- w \bm{\epsilon}_{\bm{\theta}_I}(\bar{\bx}_t, \bm{0}, t)$.
\State Apply a pre-defined sampler (\eg, DDIM~\citep{song2021denoising}) to $\bar{\bx}_{t-1}$ from $\bar{\bx}_{t}$ and $\bm{\epsilon}_t$.
\EndFor
\For{$t=T$ to $1$}
\State $\bm{\epsilon}_t = (1+w) \bm{\epsilon}_{\bm{\theta}_M}(\bz_t, \bc, \bar{\bx}_{0}, t) 
- w \bm{\epsilon}_{\bm{\theta}_M}(\bz_t, \bm{0}, \bar{\bx}_{0}, t) $.
\State Apply a pre-defined (\eg, DDIM~\citep{song2021denoising}) to $\bz_{t-1}$ from $\bz_{t}$ and $\bm{\epsilon}_t$.
\EndFor
\State Decode the clip from latents: $\bx^{1:L} = g_{\bm{\theta}}(\bar{\bx}_0,\bz_0)$.
\State Output the generated video $\bx^{1:L}$.
\end{algorithmic}
\end{spacing}
\end{algorithm}

\newpage
\section{Role of motion latent vectors}

\begin{figure}[ht!]
    \centering
    \includegraphics[width=.95\linewidth]{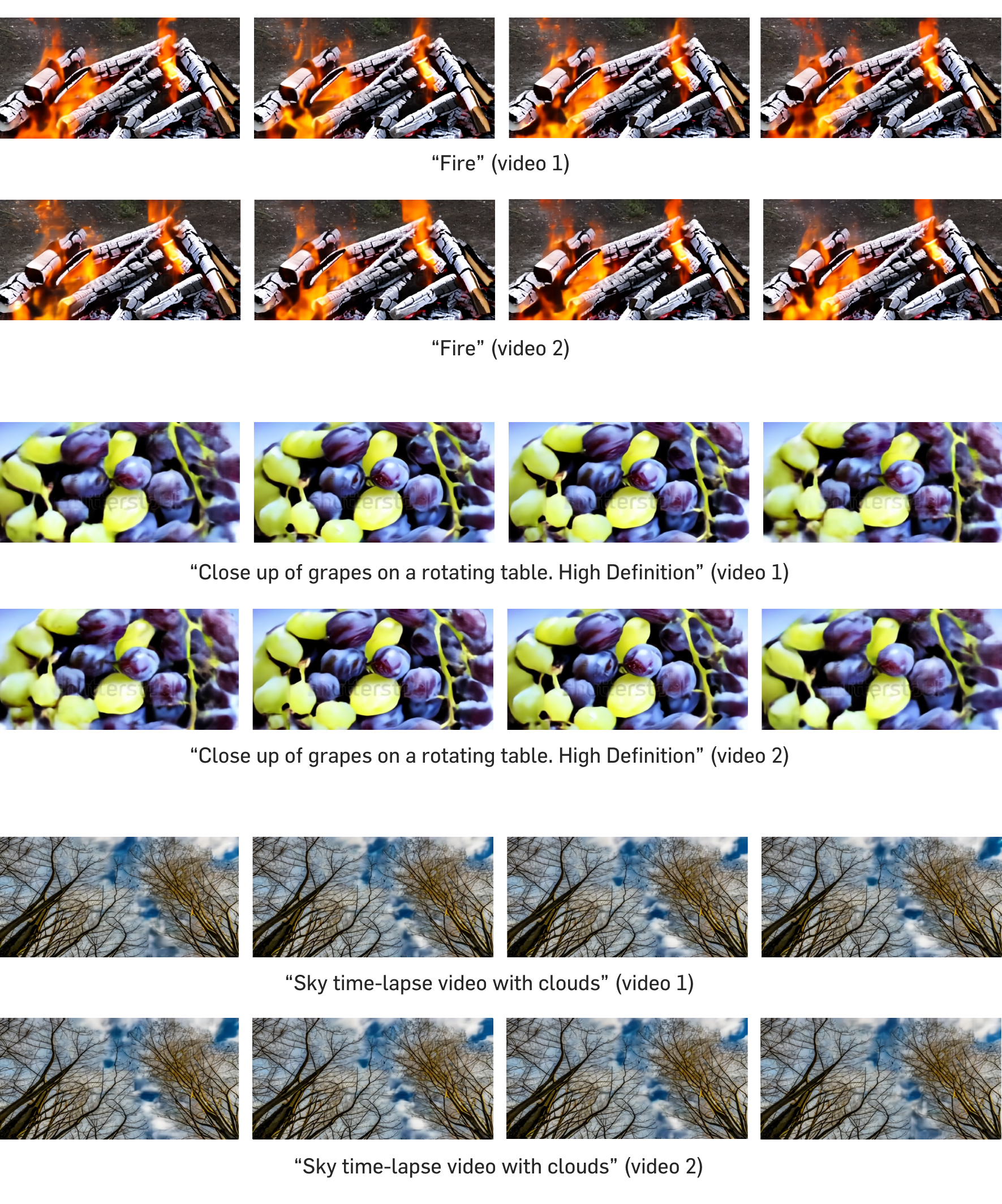}
    \caption{
    \textbf{512$\times$1024 resolution, 16-frame text-to-video generation results} from \sname, where we fix a content frame for each text prompt and sample different motion latent vectors.}
    \label{fig:motion_vector}
\end{figure}

As shown in Figure~\ref{fig:motion_vector}, given a fixed content frame, our method can generate videos with different motions, by generating different motion latent vectors.

\newpage
\section{{Quantitative results with different guidance scales}}

{In Table~\ref{tab:cfg}, we report FVD and CLIPSIM scores with different classifier-free guidance (cfg) scales.}

\begin{table}[ht!]
\captionof{table}{\textbf{T2V generation on WebVid-10M.} $\downarrow$ and $\uparrow$ indicate lower and higher scores are better, respectively.cfg denotes classifier-free guidance scale.}
\small\centering
\begin{tabular}{l c c}
\toprule
{Method} & {FVD $\downarrow$} & {CLIPSIM $\uparrow$} \\
\midrule
    {\sname (cfg=7.0)}  & {235.5} & {0.3001} \\
    {\sname (cfg=9.0)}  & {238.3} & {0.3020} \\
    {\sname (cfg=10.0)} & {245.2} & {{0.3031}} \\
    {\sname (cfg=11.0)} & {246.9} & {{0.3034}} \\
    \bottomrule
\label{tab:cfg}
\end{tabular}
\end{table}
\newpage
\section{{Qualitative results of the autoencoder}}
\begin{figure}[ht!]
    \centering
    \includegraphics[width=\linewidth]{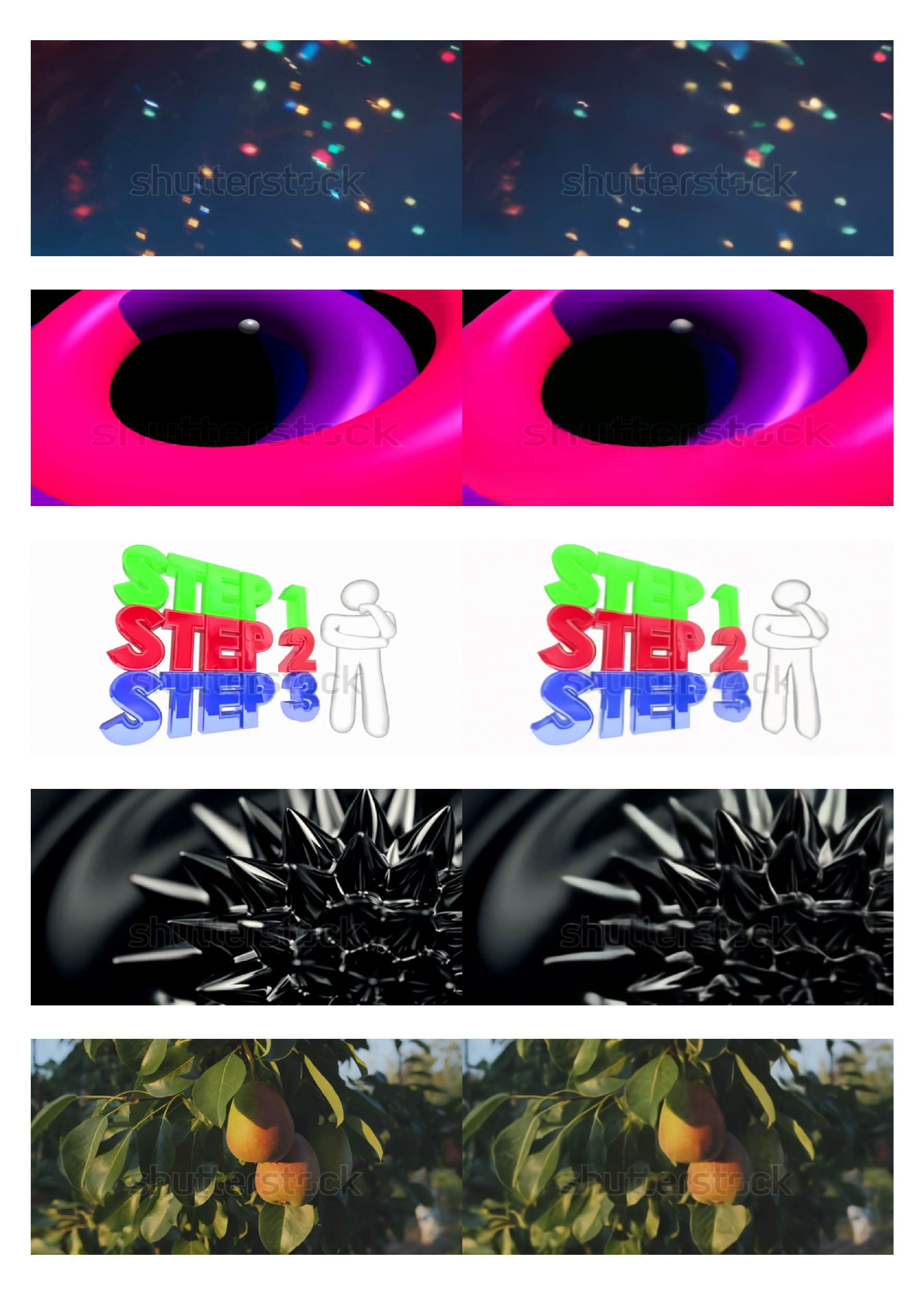}
    \caption{\textbf{512$\times$1024 resolution, 16-frame video reconstruction results} from our \sname.}
    \label{fig:recon_1}
\end{figure}

\begin{figure}[ht!]
    \centering
    \includegraphics[width=\linewidth]{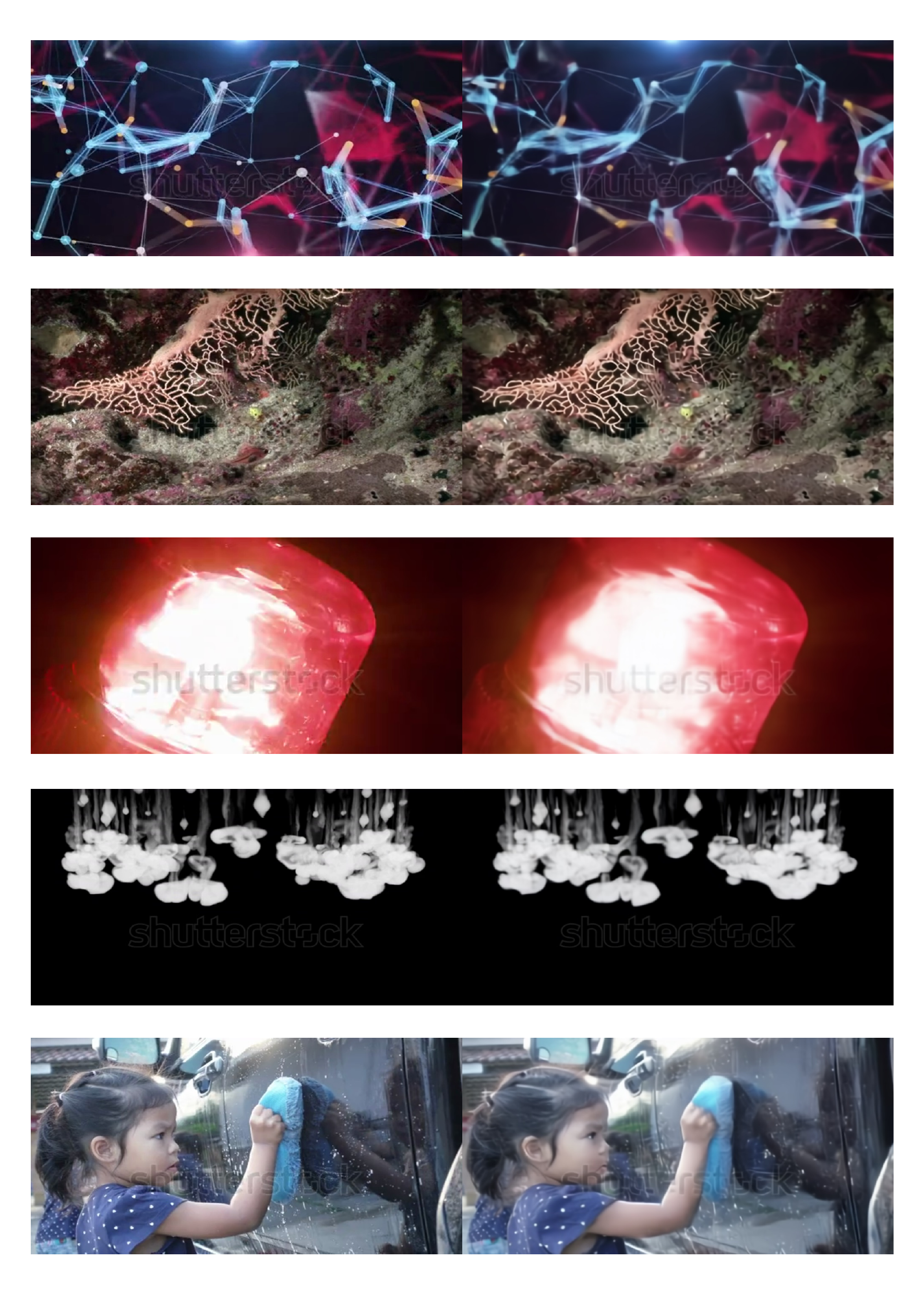}
    \caption{\textbf{512$\times$1024 resolution, 16-frame video reconstruction results} from our \sname.}
    \label{fig:recon_2}
\end{figure}

{In Figure~\ref{fig:recon_1} and \ref{fig:recon_2}, re visualize ground truth frames in the validation set in the WebVid-10M dataset (left) and reconstructions from our autoencoder (right). To visualize frames better, we only present the first frame in the paper; please refer to the supplementary material for video visualization.}

\newpage
\section{Limitation and future works}
\label{appen:future}
While \sname shows promising results on existing video generation benchmarks, there exist several limitations and corresponding future works. In what follows, we explain such limitations in addition to the limitations that we mentioned in our main text.

\textbf{Autoencoder quality.}
We found that our keyframe design works very well in pixel space, as also shown in promising results for the UCF-101 generation. While this concept also fairly worked well in latent space built in an image-wise manner (\eg, Stable Diffusion latent space~\citep{rombach2021highresolution}), we found there exists considerable frame-wise quality drop if the underlying motion in the video contains extremely dynamic motion. We hypothesize this is because the latent space that we used for reconstruction does not consider the temporal coherency of videos, resulting in less temporally coherent frames as latent vectors than frames in pixel space. We believe this limitation can be mitigated by the following solutions. First, future works can consider training our module in low-resolution pixel space first with training additional upsampler diffusion models (\ie, cascaded diffusion). Moreover, one can consider constructing the latent space from scratch using both large image and video data. Finally, our keyframe design as a weighted sum of video frames may not be an optimal choice to represent the overall contents of the video; exploring the better forms of content vectors that are similar to the original image should be an interesting future work. 

\textbf{Model size.}
We use fairly small autoencoder and diffusion models due to lack of resources used for training, compared with recent text-to-video generation models. Exploring the quality improvement with respect to the number of model parameters also should an interesting direction.

\textbf{Using negative prompts.}
We do not apply negative prompts in text-to-video generation, which have been recently used to improve the generated video quality. We believe that applying this technique to \sname will improve the video quality. 

\end{document}